\pdfoutput=1
\documentclass[11pt]{article}

\usepackage[]{acl}

\usepackage{times}
\usepackage{latexsym}
\usepackage{todonotes}

\usepackage{lipsum}  
\usepackage{booktabs}
\usepackage{adjustbox}
\usepackage{url}
\usepackage{amssymb}
\usepackage{amsmath}
\usepackage{multirow}
\usepackage{tabularx}
\usepackage{graphicx}
\usepackage{enumitem}
\usepackage{multirow}
\usepackage{mathtools}
\usepackage{tcolorbox}
\usepackage{xcolor}

\newcommand{\dataset}{\textsc{Missci}}

\usepackage[T1]{fontenc}
\usepackage[utf8]{inputenc}

\usepackage{microtype}

\title{\dataset{}: Reconstructing Fallacies in Misrepresented Science}

\author{
 Max Glockner$^\clubsuit$,
 Yufang Hou$^{\diamondsuit \clubsuit}$,
 Preslav Nakov$^\spadesuit$ \and
 Iryna Gurevych$^{\clubsuit}$
 \vspace{0.25em}\\
  $^\clubsuit$Ubiquitous Knowledge Processing Lab (UKP Lab),\\
TU Darmstadt and Hessian Center for AI (hessian.AI)\\
  $^\diamondsuit$IBM Research, Ireland,
  $^\spadesuit$MBZUAI
  \\
    \url{www.ukp.tu-darmstadt.de}
}

\begin{document}

\maketitle
\begin{abstract}

Health-related misinformation on social networks can lead to poor decision-making and real-world dangers. Such misinformation often misrepresents scientific publications and cites them as ``proof'' to gain perceived credibility. To effectively counter such claims automatically, a system must explain how the claim was falsely derived from the cited publication. Current methods for automated fact-checking or fallacy detection neglect to assess the (mis)used evidence in relation to misinformation claims, which is required to detect the mismatch between them. To address this gap, we introduce \dataset{}, a novel argumentation theoretical model for fallacious reasoning together with a new dataset for real-world misinformation detection that misrepresents biomedical publications. Unlike previous fallacy detection datasets, \dataset{} (\emph{i})~focuses on implicit fallacies between the relevant content of the cited publication and the inaccurate claim, and (\emph{ii})~requires models to verbalize the fallacious reasoning in addition to classifying it. We present \dataset{} as a dataset to test the critical reasoning abilities of large language models (LLMs), 
which are required to reconstruct real-world fallacious arguments, in a zero-shot setting. We evaluate two representative LLMs and the impact of providing different levels of detail about the fallacy classes 
to the LLMs via prompts. Our experiments and human evaluation show promising results for GPT~4, while also demonstrating the difficulty of this task.\footnote{Code and data are available at: \url{https://github.com/UKPLab/acl2024-missci}.}

\end{abstract}

\section{Introduction}
\label{sec:intro}
\begin{figure}[h!]
\small
    \centering
    \includegraphics[width=\linewidth]{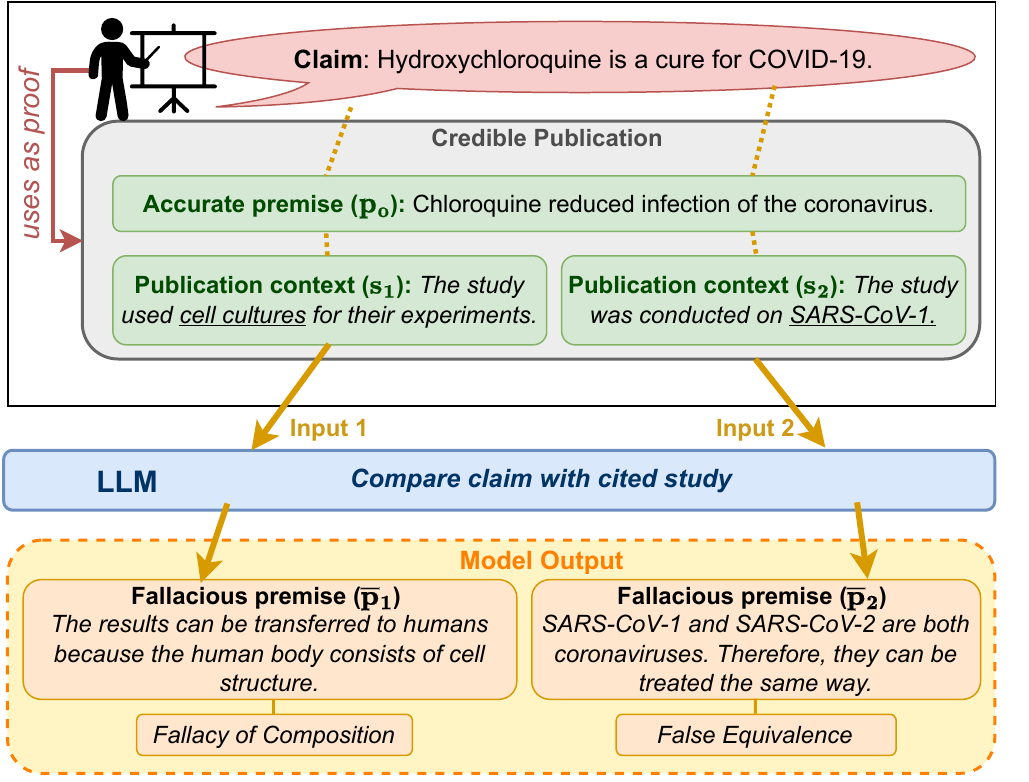}
    \caption{ 
    \textbf{Fallacious Argument Reconstruction:} 
    The claim is falsely derived from the cited study (\emph{green}) by relying on the content of $p_0$. The model generates and classifies the fallacious reasoning (\emph{orange}) that needs to be applied 
    when concluding the claim based on all relevant study content (including $s_1$ and $s_2$).
    }

    \label{fig:figure-task-main}
\end{figure}
False or misleading narratives spread rapidly on social media \citep{vosoughi2018spread,wardle20185}, posing challenges for non-experts in discerning credible information, and exceeding the capabilities of human fact-checkers (HFC).  
The need to support HFC has accelerated the research in automated fact-checking (AFC) and related tasks \citep{guo2022survey, schlichtkrull-etal-2023-intended}. 
Yet, the real-world applicability of  AFC systems is limited due to the lack of trustworthiness \citep{nakov2021automated}, or their reliance on counter-evidence, which may not exist \citep{glockner-etal-2022-missing}. 
Often, misinformation distorts genuine information rather than creating new content \citep{brennen2020types}. 
For example, the claim in Figure~\ref{fig:figure-task-main} that ``\emph{hydroxychloroquine is a cure for COVID-19}'' contains a kernel of truth and relies on some content (referred to as \emph{accurate premise)} of the cited study that found ``\emph{chloroquine reduced infection of the coronavirus}.'' 
However, further content from the study shows that it did not conduct human experiments ($s_1$) and focused on SARS-CoV-1 ($s_2$), different from the virus causing COVID-19. Only when knowing this additional information ($s_1$ and $s_2$) about the cited study, one can detect the applied fallacies (\emph{Fallacy of Composition} and \emph{False Equivalence}).

In this work, we focus on inaccurate claims that misrepresent scientific publications. We assume that the misrepresented publication is presented alongside the claim as a ``proof'' amplifying the claim's impact through increased perceived credibility. With this assumption, we can access the cited sources, which is essential for detecting fallacious reasoning and implements human verification strategies \citep{silverman2014verification}. 
Our goal is to automatically outline the fallacious reasoning and explain why the claim is incorrect, a crucial aspect of debunking misinformation \citep{schmid2019effective,lewandowsky2020debunking}.

Earlier work on fallacy detection focused on surface-level fallacies like \emph{Ad Hominem} or \emph{Loaded Language}. More recent work \citep{jin-etal-2022-logical,alhindi-etal-2022-multitask} also included logical fallacies like \emph{False Cause} or \emph{Hasty Generalization}. Yet, they did not adapt the task definition to account for the fact that these fallacies may need information beyond what is explicitly stated in the text. This hinders their applicability to real-world fallacies that rely on external information. 
To bridge this gap, we introduce \dataset{}, 
a new argumentation theoretical model for fallacious reasoning, accompanied by a new fallacy detection dataset derived from real-world misinformation.
Unlike prior work, we (\emph{i}) treat inaccurate claims as the conclusion of a logical argument, encompassing all necessary information to detect the applied fallacies (\emph{green} in Figure~\ref{fig:figure-task-main}).
We (\emph{ii})~formulate a distinct fallacy inventory drawn from literature to express fallacies when misrepresenting scientific publications. Finally, inspired by \citet{cook2018deconstructing}, we (\emph{iii})~explicitly verbalize the fallacious reasoning  via premises (\emph{orange}) that only implicitly exist based on the claim to reconstruct the fallacious argument. 

Our focus lies on the reasoning abilities to reconstruct fallacious arguments, and we manually paraphrase the relevant publication content (\emph{green}) based on HFC articles, rather than using the original misrepresented document.
Motivated by recent progress in zero-shot performance of large language models (LLMs) \citep{kojima2022large}, we evaluate the reasoning abilities of two state-of-the-art LLMs in reconstructing the fallacious reasoning on \dataset{} and define the task in a zero-shot setup, exemplified in Figure~\ref{fig:figure-task-main}.
Given the claim, an accurate premise, and the publication contexts, the model must verbalize the fallacious reasoning and assign fallacy classes. 
Our contributions are:
\begin{itemize}[noitemsep]
    
    \item A new \textbf{argumentation theoretical model} with a new task formulation to reconstruct the fallacious arguments.
    \item A new \textbf{dataset} comprising complex fallacious arguments of real-world misinformation.
    \item \textbf{Evaluation} of two state-of-the-art LLMs and their critical reasoning abilities to reconstruct fallacious arguments.
\end{itemize}

\section{Related Work}
\label{sec:related-work}
Previous work has focused on surface-level fallacies for propaganda detection \citep{da-san-martino-etal-2019-fine,piskorski-etal-2023-multilingual,salman2023detecting}, for online discussions \citep{habernal-etal-2018-name,sahai-etal-2021-breaking}, or for gamified settings \citep{habernal-etal-2017-argotario}. The addressed fallacies typically do not require information beyond what is stated explicitly in the text. Other work targeted specific fallacies such as \emph{Non Sequitur} in law \citep{nakpih2020automated} or \emph{Ad Hominem} in social media \citep{habernal-etal-2018-name}. More similar to our work, some research \citep{jin-etal-2022-logical,musi2022developing,alhindi-etal-2022-multitask} focused on logical fallacies from the real world. Yet, they neither verbalized the implicitly applied fallacies, nor considered the underlying sources beyond what is explicitly stated in the text. Moreover, our task design to generate fallacious premises differs from implicit premise detection work \citep{habernal-etal-2018-argument, chakrabarty-etal-2021-implicit} in that the  premises in \dataset{} are inherently invalid and linked to applied fallacy classes (see $\overline{p}_1$ and $\overline{p}_2$ in Figure~\ref{fig:figure-task-main}; \emph{orange}). Existing work on fallacy generation focused on data generation \citep{huang-etal-2023-faking, alhindi2023large} rather than on articulating applied fallacious reasoning within real-world fallacious arguments.

Scientific AFC work \citep{wadden-etal-2020-fact,sarrouti-etal-2021-evidence-based,wadden-etal-2022-scifact,lu2023scitab,vladika-matthes-2023-scientific} considered external sources, like us, but did not identify and articulate the nuanced fallacies when concluding a claim from the cited study. Detecting distortions in scientific communication is part of science communication research \citep{augenstein-2021-determining}, where studies have examined exaggerations in news articles \citep{sumner2014association,bratton2019association,yu-etal-2020-measuring,wright-augenstein-2021-semi}, analyzed reporting certainty in scientific publications \citep{pei-jurgens-2021-measuring}, or quantified information mismatches between reported and actual scientific findings \citep{wright-etal-2022-modeling}. In parallel work, \citet{wuhrl2024understanding} quantify and compare the original paper with other reporting of their findings across fine-grained dimensions such as certainty or generalization. In addition to our distinct task setup, our problem space differs as we focus on harmful misinformation that comprises more severe distortions, which are not necessarily bound to a study's main findings.

\section{Formalism of \dataset{}}
\label{sec:preliminaries}
\subsection{Misrepresented Science Arguments}
\label{sec:preliminaries:fallacious-argument}
\begin{figure}
\small
    \centering
    \includegraphics[width=\linewidth]{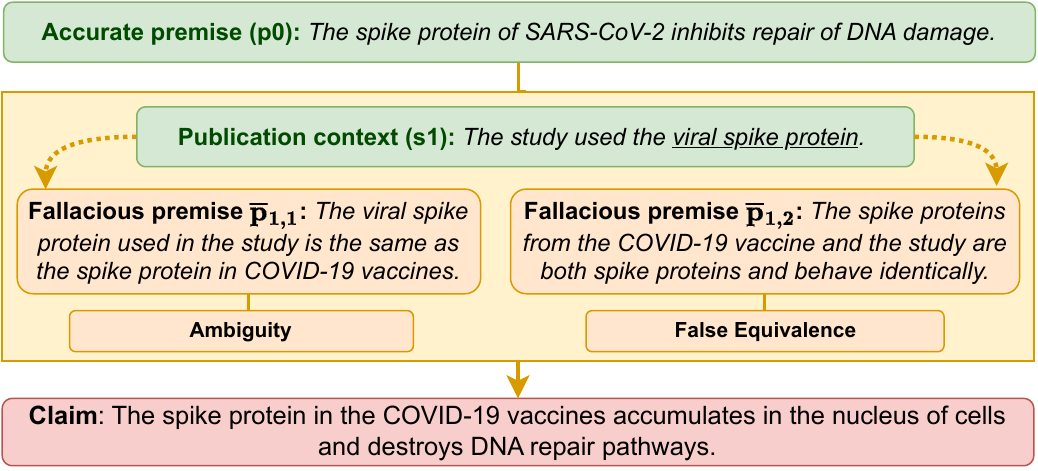}
    \caption{
    \textbf{Interchangeable Fallacies:}
On the left,  no distinction between the different ``spike proteins'' (from the vaccine or the virus) is made;
on the right, both
are assumed to behave alike.  
Only one of these premises is needed to bridge the reasoning gap.
}
    
    \label{fig:figure-alternative-fallacies-example}
\end{figure}

Inspired by \citet{cook2018deconstructing}, we reconstruct the fallacious reasoning in the form of a logical argument. For an accurate claim $c$, a logical argument would be structured as follows:
\begin{equation}
    \{p_0\,,\,  p_1\,,\,  \ldots \,,\,  p_N \} \Rightarrow c
\end{equation}
where $P=\{p_0, p_1, \ldots, p_N\}$ is a set of \emph{accurate} premises that jointly entail ($\Rightarrow$) the true claim $c$.
For inaccurate claims $\overline{c}$, the entailment relation does not hold based on accurate premises, formulated as $P \not\Rightarrow \overline{c}$, where $\not\Rightarrow$ denotes a corrupted entailment relation. 
To \emph{reconstruct} a fallacious argument, a set of inaccurate (fallacious) premises $\overline{p}_i \in \overline{P}$ must be applied such that $P \cup \overline{P} \Rightarrow \overline{c}$. 
For example, consider the following argument (simplified from Figure~\ref{fig:figure-alternative-fallacies-example}):
\begin{itemize}[noitemsep]
    \item \textbf{Accurate premise $p_0$:} The viral COVID-19 spike protein inhibits repair of DNA damage.
    \item \textbf{Fallacious premise $\overline{p}_{1,2}$:} COVID-19 vaccine spike proteins are as dangerous as viral COVID-19 spike proteins.
    \item \textbf{Conclusion ($\overline{c}$):} Therefore, COVID-19 vaccines inhibit repair of DNA damage.
\end{itemize}
Here, the accurate premise alone lacks sufficient support for the conclusion (or claim). Establishing a support relationship between the accurate premise and the conclusion requires the fallacious premise, which employs the \emph{False Equivalence} fallacy.
To debunk a claim $\overline{c}$, the argument can be \emph{deconstructed} by highlighting the fallacies applied in each $\overline{p}_i$. This invalidates the premises that are necessary to establish the claim as a logical conclusion and renders the argument invalid.

Misinformation that is falsely derived from a single credible publication $S$ can be formulated as $S \not\Rightarrow \overline{c}$.
Each argument in \dataset{} has exactly one accurate premise, $P = \{p_0\}$, which is entailed by (parts of) $S$ and represents the ``kernel of truth'' behind the inaccurate claim $\overline{c}$. 
Since all arguments in \dataset{} constitute misinformation, each argument is illogical and contains at least one reasoning gap, which 
must be bridged via a fallacious premise $\overline{p}_i$ that applies a fallacy class $f_i$ to support the claim. This reasoning gap only becomes imminent after observing relevant context ($s_i$) from the misrepresented publication $S$. For example, consider Figure~\ref{fig:figure-alternative-fallacies-example}, where a study observed harm from SARS-CoV-2 spike proteins (\emph{accurate premise $p_0$}). The argument wrongly applied this finding to mRNA vaccines, assuming that viral and vaccine spike proteins behave identically (in both (alternative) fallacious premises). Without detailed content about the misrepresented publication (that it was observed on \emph{viral} spike proteins in $s_1$), this fallacy is undetectable.
In this work, we manually paraphrase this necessary context $s_i$ that exhibits the reasoning gap from the publication $S$ where each $s_i$ is entailed by $S$.
Because  the accurate premise $p_0$ is always entailed by the publication, 
its publication context ($s_0$) is identical to the accurate premise. 
We represent each fallacious reasoning $R_i \in R$ that bridges one reasoning gap as a triplet $R_i = $ ($s_i$, $\overline{p}_i$, $f_i$). To bridge all reasoning gaps and fully support the claim $\overline{c}$ each fallacious reasoning $R_i \in R$ must be applied.
The overall formulation of a fallacious argument $\mathcal{A}$ is shown below:
\begin{equation}
\biggl\{\,\, 
\overset{\substack{s_0 \\ =}}{p_0} 
\,,\, 
\underset{\substack{\downarrow \\ f_1}}{\overset{\substack{s_1 \\ \downarrow}}{\overline{p}_1}}
\,,\, 
\underset{\substack{\downarrow \\ f_2}}{\overset{\substack{s_2 \\ \downarrow}}{\overline{p}_2}} 
\,,\, 
\ldots
\,,\, 
\underset{\substack{\downarrow \\ f_N}}{\overset{\substack{s_N \\ \downarrow}}{\overline{p}_N}} 
\,\, 
\biggr\}\,\, 
\Rightarrow \overline{c}
\end{equation}
Each argument $\mathcal{A} = (\overline{c}, p_0, R)$ comprises an inaccurate claim $\overline{c}$, that builds on the accurate premise $p_0$ 
and applies at least one fallacious reasoning $R_i \in R$.
As shown in  Figure~\ref{fig:figure-alternative-fallacies-example}, in some cases, multiple fallacious premises with distinct fallacy classes can be used interchangeably (i.e., $\overline{p}_i = \bigl[\overline{p}_{i,1},\:\overline{p}_{i, 2}\bigr]$). Interchangeable fallacies always share the identical publication context ($s_i$), but only one of them is necessary to bridge the reasoning gap.

\subsection{Non-exhaustive, Inductive Arguments}
\label{sec:preliminaries:inductive}
Unlike \citet{cook2018deconstructing}, we do not require arguments to deduce the claim, which is unrealistic based on empirical evidence from scientific publications. Instead, we consider \emph{strong inductive support} sufficient for ($\Rightarrow$).  
In inductive arguments, invalid premises, by definition, weaken the conclusion without necessarily rendering it false. To avoid labeling minor mismatches as fallacies, we ensure the relevance of each fallacious reasoning $R_i \in R$ in the strong inductive argument by deriving $R$ exclusively from the HFC article. 
By relying on the HFC, the extracted fallacious reasoning lines are non-exhaustive and limited to the most pivotal ones.
Importantly, fallacies within the logical arguments do not necessarily match the fallacies made by the claimant. For example, one can make the claim in Figure~\ref{fig:figure-task-main} after only skimming parts of the study without ever knowing that experiments were performed in cell cultures ($s_1$). In this case, the \emph{Fallacy of Exclusion} rather than the \emph{Fallacy of Composition} was applied. To account for these cases, we state our objective as detecting the necessary fallacies needed to conclude the claim $\overline{c}$ from all relevant content of the misrepresented publication $S$. This follows the \emph{principle of total evidence}, which dictates that an inductive argument must consider all available relevant evidence \citep{chakraborti2007logic}.

\subsection{Task Definition}
\label{sec:preliminaries:task-definition}
We assess the ability to reconstruct the fallacious reasoning for each fallacious argument $\mathcal{A}$ on the fallacious reasoning level: For each fallacious reasoning $R_i \in R$, given the claim $\overline{c}$, the accurate premise $p_0$ and the publication context $s_i$ from $R_i$, the model must verbalise the fallacious premise $\hat{\overline{p}}_i$ and predict the applied fallacy class $\hat{f_i}$ to bridge the reasoning gap, so that ($\hat{\overline{p}}_i$, $\hat{f_i}$) constitute valid fallacies as approximated via the annotated interchangeable fallacies ($\overline{p}_i$, $f_i$) of $R_i$. % in \dataset{}.

\section{Dataset}
\label{sec:dataset}

\begin{table}[]
\small
    \centering
    \begin{tabular}{lccc}
    \toprule
    & \textbf{Collect} & \textbf{Select} & \textbf{Reconstruct}  \\
    \midrule
    \textbf{HFC articles} & 527 & 150 & 147  \\
    \textbf{Links} & 8,695 & 208 & 184 \\
    \midrule
    \textbf{Arguments} & -- & -- & 184  \\
    \textbf{Fall. Reasoning $\mathbf{R_i}$}  & -- & -- & 435  \\

     \bottomrule
     
    \end{tabular}
    \caption{\textbf{Dataset Construction:}
     Number of elements for all three steps during dataset construction. 
     }
    \label{tab:dataset-collection-overview}
\end{table} 

Our main annotator was a M.Sc. student in biology, covering early pilot studies and post-annotation consolidation. Additionally, two more M.Sc. students, one in biology and one in linguistics, were employed during annotation. The annotators received a pay of 12.26 EUR per hour. We used Surge AI\footnote{\url{https://www.surgehq.ai/}}
as the annotation tool. Weekly meetings involving all annotators and one of the authors were held throughout the project to provide feedback and to refine the guidelines as needed, in line with the recommendations of \citet{klie2023analyzing}.
To create \dataset{}, we (1) \emph{collected} HFC articles and pre-selected links that may point to a misrepresented publication, (2) manually \emph{selected} all links that pointed to a misrepresented publication, and (3) \emph{reconstructed} the fallacious arguments from the HFC articles.
A summary of these three steps is given in  Table~\ref{tab:dataset-collection-overview}. We collected a total of 527 fact-checking articles from HealthFeedback\footnote{\url{https://healthfeedback.org/}} until January 2023, excluding those that address accurate claims. HealthFeedback collaborates with scientists in reviewing health and medical claims. From these HFC articles, we annotated 8,695 links from reputable sources (cf. §\ref{appendix:dataset:step1:filtering}) to %if
determine whether a link pointed to a misrepresented scientific publication.
Our annotators found 208 links pointing to misrepresented scientific publications across 150 HFC articles (cf. §\ref{appendix:dataset:step1:process}; Krippendorff's $\alpha$ was $0.728$).

\subsection{Fallacious Argument Reconstruction}
\label{sec:data:step2}

The annotators were instructed to generate all elements of the fallacious argument $\mathcal{A}$ that falsely concludes the claim $\overline{c}$. 
This included the accurate premise $p_0$ as well as the fallacious premise $\overline{p}_i$, fallacy class $f_i$, and publication context $s_i$ for each fallacious reasoning $R_i$ (cf. §\ref{appendix:dataset:fallacies} for a list of all fallacy classes).
Each element had to be justified with an extracted statement from the HFC article. 
Often, selecting a single definitive fallacious reasoning was ambiguous and oxymoronic, akin to identifying the ``correct invalid reasoning'' (cf. §\ref{sec:preliminaries}; Figure~\ref{fig:figure-alternative-fallacies-example}). This aligns with \citet{bonial-etal-2022-search}, who observed that due to overlapping definitions, fallacies could often be reformulated to fit the definition of a different fallacy. Hence, we allowed separate listing of interchangeable fallacies, which were merged during consolidation.
As we aimed to detect the fallacious reasoning \emph{between} a scientific publication and an inaccurate claim, our work rests on the assumption that the publication itself is trustworthy. To verify the trustworthiness, the annotators rated the credibility of the scientific document based on the HFC article, which we analyzed in §\ref{appendix:publication-credibility}.
Detailed instructions and the annotation process are outlined in §\ref{appendix:dataset:argument-construction}.

\subsection{Inter-Annotator Agreement}
We collected 520 annotated HITs for 208 potential arguments.
After consolidation (cf. §\ref{appendix:dataset:consolidation}), \dataset{} contained 435 distinct fallacious reasoning lines ($R_i$) bridging different reasoning gaps (with a total of 550 interchangeable fallacies) for 184 fallacious arguments. Each argument involved 1-5 fallacious reasoning lines ($R_i$), averaging at 2.4 per argument. Most of the arguments within \dataset{} were related to the COVID-19 infodemic. We show the distribution of arguments over years and their relation to COVID-19 in §\ref{appendix: claim-topics}.
Calculating the inter-annotator agreement for the fallacy class annotations faced two challenges: interchangeable fallacies with different but valid labels, and annotators identifying different (non-interchangeable) fallacious reasoning lines that bridge different reasoning gaps. 

To address this, we used two complementary measures:
We calculated the inter-annotator agreement for the fallacy class $f_i$  among all 253 fallacious reasoning $R_i$ identified by at least two annotators within the consolidated arguments. 
When simulating a single-label classification setup, the inter-annotator agreement, measured using Krippendorff's $\alpha$, was 0.520. This is comparable to Cohen's $\kappa$ of 0.47 \citep{alhindi-etal-2022-multitask} and 0.52 \citep{musi2022developing} in similar work.\footnote{Our agreement measure differs as we employed up to three rather than two annotators.}
We additionally compared each fallacious reasoning $R_i$ identified by each individual annotator to the consolidated argument and measured how many fallacious reasoning $R_i$ of the consolidated argument a single annotator found on average. 
Here, we considered all fallacies that were merged during consolidation as identical, and did not differentiate whether the annotators selected distinct interchangeable fallacies that apply different fallacy classes.
Here, we only considered 70 arguments, which were fully annotated by all three annotators, for the computation to not artificially inflate the coverage by a single annotator. On average, each annotator identified 72.5\% of the fallacious reasoning lines $R_i$ in the consolidated argument. We examined how this affected the overall recall of the detected fallacies in \dataset{} in §\ref{sec:fallacy-coverage}.

%\section{Analysis of \dataset{}}
\label{sec:dataset-analysis}

\subsection{Fallacy Class Analysis}
\label{sec:interchangable-fallacy-analysis}
\begin{figure}[h]
\small
    \centering
    \includegraphics[width=\linewidth]{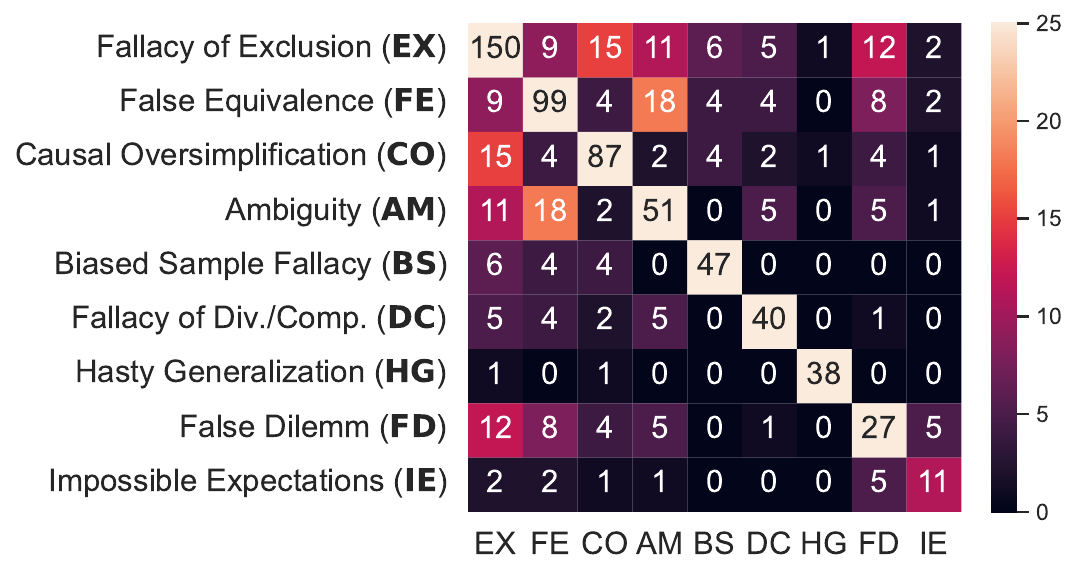}
    \caption{
    \textbf{Interchangeable Fallacy Classes:} 
    Heatmap of co-occurring interchangeable fallacy classes of the consolidated arguments ordered by frequency.
    }  
    \label{fig:fallacy-confusion}
\end{figure}

To understand which fallacy classes have been annotated together as interchangeable fallacies, we show their co-occurrence matrix in Figure~\ref{fig:fallacy-confusion}. 
The definitions and the examples for all fallacies are given in §\ref{appendix:dataset:fallacies}. We observe that most co-occurrences are between \emph{False Equivalence} and \emph{Ambiguity} (as discussed in Figure~\ref{fig:figure-alternative-fallacies-example}). In contrast, \emph{Hasty Generalization} was the most clear-cut fallacy in our annotations, likely because HFC typically explicitly specify when a study lacks sufficient observations for the claim, and because it had little overlap with other fallacy definitions.
The majority class of all fallacy class annotations is the  \emph{Fallacy of Exclusion}. 
This fallacy omits critical information when drawing a conclusion and could theoretically apply to every reasoning gap that depends on the information in the publication context $s_i$, because each $s_i$  contains content that undermines the claim. 
To address this, during annotation the annotators were tasked to prioritize more specific fallacies before the \emph{Fallacy of Exclusion} (cf. §\ref{appendix:dataset:fallacies-selection-priorities}), yet the fallacy class remains the most common.

We found two main reasons for the prevalence of the \emph{Fallacy of Exclusion} in \dataset{}:
The first and primary reason for including this fallacy in our inventory, is that parts of the misrepresented publication contradict the claim $\overline{c}$. For example, the claim that ``\emph{spike proteins induced by RNA vaccines can damage blood vessels}'' is based on a study 
which concludes that ``\emph{vaccination could protect against blood vessel damage}''. In this case, the authors' comment must be ignored and no other fallacy in our inventory can be used to conclude a claim opposite to their conclusion.
Second, the \emph{Fallacy of Exclusion} often serves as a fallback class in \dataset{} due to its broad applicability. Given the inevitability of an incomplete fallacy inventory, instances where the detected fallacies do not clearly align with the predefined fallacy classes are frequently labeled as \emph{Fallacy of Exclusion}. This leads to co-occurrences with other fallacies in borderline cases. For example, the claim that ``\emph{Pfizer's COVID-19 vaccine effectiveness dropped from 100\% to 20\%}'' relies on infection numbers, and ignores the reported high effectiveness against severe disease. This flawed reasoning could be interpreted as \emph{False Equivalence}, assuming mild and severe COVID-19 cases are equivalent, or \emph{Fallacy of Exclusion} by omitting the protection against severe diseases. A clear fallacy class can only be assigned for a specific \emph{verbalized} fallacious premise.

\section{Experiments}
\label{sec:experiments}

\begin{table*}[]
\small
    \centering
    \begin{tabular}{l|ccc | cccc | c}
    \toprule
    & \multicolumn{3}{c|}{\emph{Fallacy}} & \multicolumn{4}{c|}{\emph{Fallacious premise (@1)}}  & \emph{Consistency} \\
    \textbf{Model} & \textbf{P@1} & \textbf{Arg@1} & \textbf{F1} (\emph{micro}) & \textbf{METEOR} & \textbf{BERTScore} & \textbf{NLI-A} & \textbf{NLI-S} & \textbf{Matches@1 (\%)}\\
    \midrule
    \emph{random} + \emph{claim} & \multirow{2}{*}{0.131} & \multirow{2}{*}{0.264} & \multirow{2}{*}{0.117} & 0.181 & 0.611 & 0.120 & 0.130 & --\\
    \emph{random} + $p_0$ &  & &  & 0.188 & 0.599 & 0.062 & 0.067 & --\\

    \midrule
LLaMA~2   (D) & 0.223 & 0.416 & 0.233 & 0.222 & 0.617 & 0.123 & 0.148 & 40.5\\
LLaMA~2   (DE) & 0.209 & 0.422 & 0.232 & 0.229 & 0.621 & 0.124 & 0.148 & 34.7\\
LLaMA~2   (DL) & 0.196 & 0.409 & 0.211 & 0.203 & 0.616 & 0.130 & 0.143 & 41.0\\
LLaMA~2   (DLE) & 0.209 & 0.416 & 0.233 & 0.207 & 0.616 & 0.129 & 0.145 & 19.3\\
LLaMA~2   (L) & 0.193 & 0.377 & 0.208 & \textbf{0.253} &  \textbf{0.627} & \textbf{0.140} & \textbf{0.165} & 54.5\\
    LLaMA~2  (LE)  & 0.212 & 0.409 & 0.222 & 0.180 & 0.609 & 0.121 & 0.134 & 43.0\\
    \midrule
    GPT~4 (D) & \textbf{0.317} & \textbf{0.571} & \textbf{0.297} & 0.239 & 0.619 & 0.069 & 0.126 & \textbf{61.2}\\
    GPT~4 (L) & 0.292 & 0.526 & 0.290 & 0.238 & 0.613 & 0.064 & 0.140 & \textbf{61.4}\\
    
    \bottomrule

    \end{tabular}
    \caption{\textbf{Argument Reconstruction Results:} Evaluation of LLaMA~2 (70B) and GPT~4 over the predicted fallacy class and the generated fallacious reasoning. We report the performance when using prompts with fallacy \emph{(D)}efinitions, \emph{(L)}ogical forms and/or \emph{(E)}xamples, and provide a consistency estimate of the LLM by asking each LLM to separately classify the fallacy present in the generated premise.}
    \label{tab:subtask3-contextwise-generation}
\end{table*} 
For each input ($\overline{c}$, $p_0$, $s_i$), comprising the incorrect claim, the accurate premise, and the publication context linked to a fallacy, the model must generate at least one fallacious premise $\hat{\overline{p}}_i$ together with the applied fallacy class $\hat{f_i}$. 
We only experiment in a zero-shot setting, since the dataset construction depends on high-quality HFC articles, which limits size and scalability. However, we separate 30 arguments
as a validation split\footnote{No misrepresented publication occurs in \emph{test} and \emph{dev}.} to allow for a prompt selection without compromising the evaluation on the unseen test split, which comprises the remaining 154 arguments with 363 fallacious reasoning lines $R_i$ bridging different reasoning gaps.

\subsection{Metrics}
Even though multiple interchangeable fallacies may be applicable, only one of them is required to reconstruct the fallacious argument.
Hence, to evaluate the fallacy classes, we report  P@1  as our primary metric, where the top-ranked predicted fallacy class $\hat{f}_{i, 1}$ is considered correct if it matches any gold fallacy class $f_{i,j}$ of the interchangeable fallacy classes in $R_i$. 
Further, we model the fallacy classification as a multi-label, multi-class classification problem, in which we ask the model to identify \emph{all} interchangeable fallacy classes within each $R_i$. While the single-label classification measures sufficiency, the multi-label multi-class classification relates to the comprehensiveness of the detected fallacy classes. Akin to previous fallacy detection work \citep{dimitrov-etal-2021-detecting, jin-etal-2022-logical} with high class-imbalances, we report the micro F1-score.
Additionally, we assume that correctly detecting at least one fallacy is sufficient to reject the claim, and report argument-level accuracy, denoted as \emph{Arg@1}, by considering an argument as rejected if the top-ranked fallacy class prediction of any fallacious reasoning $R_i$ is correct.

To evaluate the generated fallacious premises, we first match the top-ranked generated premises with the gold premises as reference texts via the predicted fallacy class and cosine similarity (cf. §\ref{appendix:experiments:metrics}). We then
report METEOR score~\citep{banarjee2005}, which was used 
for rationales in the real-world AFC dataset AVeriTeC~\citep{schlichtkrull2023averitec}, and BERTScore~\citep{bert-score} to account for semantic similarity.
Further, we follow \citet{honovich-etal-2022-true} who use a T5~\citep{raffel2020exploring} model trained on NLI data and consider the predicted probability for the entailment label as measure. Rather than using the entailment probability given the reference premise $\overline{p}_i$ as \emph{premise} and the generated  premise $\hat{\overline{p}}_i$ as \emph{hypothesis}  $e(\overline{p}_i, \hat{\overline{p}}_i)$ (denoted as \emph{NLI-A}), we additionally compute a symmetric variant (\emph{NLI-S}) via $\mathrm{max}\bigl[e(\overline{p}_i, \hat{\overline{p}}_i); e(\hat{\overline{p}}_i, \overline{p}_i)\bigr]$ to not penalize a model 
if the generated premise is more specific than the reference premise. 
More details about the metrics and matching with reference text are provided in  §\ref{appendix:experiments:metrics}. 
Finally, we measure the LLM's internal consistency of the generated premise and fallacy class, by prompting the same LLM again to classify the fallacy present in the generated fallacious premise $\hat{\overline{p}}_i$ given ($\overline{c}$, $p_0$, $\hat{\overline{p}}_i$). We report the percentage in which the same fallacy class is predicted.

\subsection{Models}
We evaluated two baselines that predict a randomly selected fallacy class. For the fallacious premise, these baselines either always predict the claim $\overline{c}$ or the accurate premise $p_0$, both of which are topically related to the gold fallacious premise but meaningless in their verbalized reasoning.

We conducted experiments with two state-of-the-art LLMs: LLaMA 2 (70B) \citep{touvron2023llama} as an open-source LLM which can be run on a local machine, and GPT~4 \citep{openai2023gpt4} as a proprietary LLM. In line with our annotation process, we prompted the LLM to generate a ranked list of multiple pairs consisting of the fallacious premise and fallacy class ($\hat{\overline{p}}_{i,j}$, $\hat{f}_{i,j}$), which may express interchangeable fallacy classes. 
We evaluated different prompts, varying in the amount of information provided to the LLMs about the fallacy classes. Specifically, we examined the impact of fallacy definitions (\emph{D}), the logical form (\emph{L}), and the examples (\emph{E}) from our fallacy inventory (cf. §\ref{appendix:dataset:fallacies}), sourced from \citet{bennett2012logically} and \citet{cook2018deconstructing}. 
The definitions offer descriptive information about the fallacies.
The logical forms abstract from the content, but explicitly indicate the applied fallacious reasoning. For instance, the logical form for the \emph{Fallacy of Composition} is \emph{``A is part of B. A has property X. Therefore, B has property X.''}. This resembles surface patterns that were found to be beneficial in logical fallacies \citep{jin-etal-2022-logical}. We hypothesize that different types of information have varying effects on fallacy classification and fallacious premise generation. We selected the best prompt based on the P@1 performance on the validation split (cf. §\ref{appendix:st3:argument-reconstructuion}).
For GPT~4, we only report the results based on the respective best LLaMA~2 prompts for fallacy premise generation and fallacy class P@1 on the test set for comparison. Prompts and hyper-parameters are in §\ref{appendix:reprocudibility}.

\subsection{Argument Reconstruction Results}
\label{sec:experiments:main-results}

Table~\ref{tab:subtask3-contextwise-generation} shows the results for reconstructing the fallacious argument. LLaMA~2 achieves its best fallacy detection (P@1) and fallacious premise generation performance using (\emph{D}) or (\emph{L}) in the prompt, respectively, which is consequently reported for GPT~4. For both LLMs, using only the fallacy definition leads to the best fallacy classification performance. Here, GPT~4 outperforms LLaMA~2 by a large margin, correctly identifying at least one fallacy in 57\% of the arguments. 
 For fallacious premise generation, each LLM exhibits the best performance based on different prompts. 
In the fallacious premise generation, even GPT~4 achieves low scores, particularly compared to the random baselines. The generated premises perform primarily poorly when the predicted fallacy class does not match the gold fallacy class of the reference premise. 
When separately evaluating the generated fallacious premises over correctly classified fallacies with matching classes only (cf. §\ref{appendix:experiments:premise-eval-correct-incorrect}; Table~\ref{tab:subtask3-contextwise-generation-correct-incorrect}),  GPT~4 surpasses the random baseline and outperforms LLaMA~2 in METEOR ($0.264$ vs. $0.243$) and BERTScore ($0.637$ vs. $0.622$), reaching comparable performance in NLI-S ($0.267$ vs. $0.266$).
 Finally, both models seem to show (slightly) improved consistency (last column) when given the logical form, suggesting that it helps the fallacy generation to match the expected form. Overall, the consistency is much higher for GPT~4.

\subsection{Fallacy Classification Results}
\label{sec:experiments:fallacy-classification}

\begin{table}[]
\small
    \centering
    \begin{tabular}{ll|cc}
    \toprule
    \textbf{LLM} & \textbf{Prompt} & \textbf{Acc.} & \textbf{F1} \\
    \midrule
    & -- &  0.493 & 0.406 \\
     & Def. &  0.577 & 0.464 \\
    & Def. + Logical &  0.630 & 0.476 \\
    LLaMA~2& Def. + Example &  0.637 & 0.476 \\
    & Def. + Logical + Example & 0.568 & 0.459 \\
    & Logical & 0.601 & 0.472 \\
    & Logical + Example & \underline{0.645} & \underline{0.499} \\

    \midrule
    & Def. &  0.738 & 0.649\\
    GPT~4 & Logical  & 0.744 & 0.624\\
     & Logical + Example & \textbf{0.771} & \textbf{0.682} \\
    
    \bottomrule

    \end{tabular}
    \caption{\textbf{Fallacy Classification:} Performance when predicting the gold fallacy class $f_{i,j}$ given the claim $\overline{c}$, the fallacy context $s_i$ and the verbalized fallacious premise $\overline{p}_{i,j}$. We report accuracy and F1-score (macro).}
    \label{tab:subtask3-fallacywise-ablation}
\end{table} 
\begin{figure}[b]
\small
    \centering
    \includegraphics[width=\linewidth]{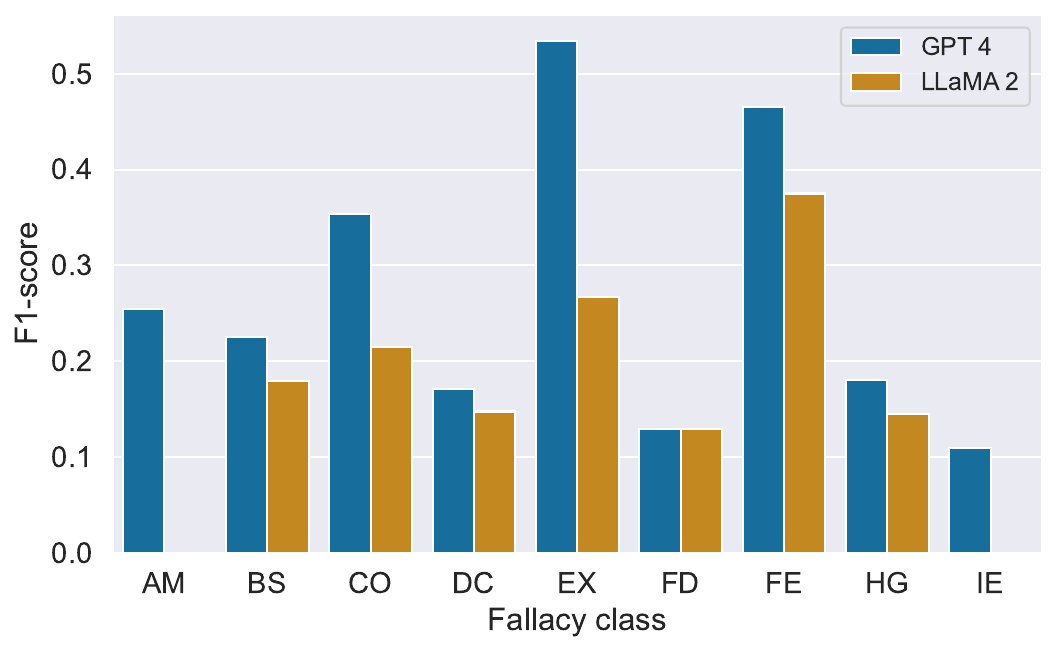}
    \caption{
    \textbf{Performance per Fallacy:}
    F1-score per predicted fallacy class from a multi-label multi-class perspective considering all model predictions.}  
    \label{fig:analysis:model-comparison-per-fallacy-with-context}
\end{figure}

We instructed LLMs to classify the applied fallacy class $f_{i,j}$ based on the \emph{provided} gold fallacious premise $\overline{p}_{i,j}$, along with the claim $\overline{c}$, the accurate premise $p_0$, and the publication context linked to a fallacy $s_i$, and report the results in Table~\ref{tab:subtask3-fallacywise-ablation}. Since each fallacious premise $p_{i,j}$ verbalizes a single fallacy class $f_{i,j}$ this becomes a single-label classification problem.
These experiments help to (\emph{i})~compare the difficulty of detecting fallacies with \emph{explicit} fallacious reasoning provided or not (as in §\ref{sec:experiments:main-results}), and (\emph{ii})~re-evaluate the LLMs and the prompts used to assess the consistency over the gold fallacious premises.
In addition to exploring the impact of  (\emph{D}, \emph{L}, \emph{E}), we also evaluated the performance when only provided with the fallacy names (first row), to assess whether sufficient fallacy knowledge was acquired during pretraining. For GPT~4, we evaluated the best prompt based on LLaMA~2 performance, as well as the prompts used to measure consistency in Table~\ref{tab:subtask3-contextwise-generation}.
Both LLMs performed strong across all prompts, especially considering that this is a 9-way classification problem.
 In §\ref{appendix:analysis:implicit-classification}, we further provide empirical evidence that GPT~4 benefits from the premise generation task, when no gold fallacious premise is available.
The accuracy over the gold premises always exceeded the consistency scores, 
suggesting that model-generated premises were not as clear-cut to a single fallacy class compared to gold fallacious premises, even by the LLMs' own judgement.
The primary misclassification for both LLMs occured between  \emph{Ambiguity} and \emph{False Equivalence} (cf. §\ref{appendix:st3:fallacy-classification}), two very related fallacies (cf. Figures~\ref{fig:figure-alternative-fallacies-example} \& \ref{fig:fallacy-confusion}). However, LLaMA~2 overpredicted \emph{False Equivalence} in general.
The best performance was reached with access to the logical form and the examples.
We hypothesize that both were most influential to our annotators, and are hence helpful for detecting their generated fallacies.

\section{Analysis}
\label{sec:analysis}
 
\subsection{Fallacy-level Performance}
We assess the performances per fallacy class for the best prompts (\emph{D}) in §\ref{sec:experiments:main-results} for both LLMs in Figure~\ref{fig:analysis:model-comparison-per-fallacy-with-context}. Specifically, we report the fallacy-level F1-score in a multi-label multi-class setting. GPT~4 outperforms LLaMA~2 in almost all classes. The strongest F1-score by LLaMA~2 is achieved for the \emph{False Equivalence} class. This aligns with the (\emph{LE}) prompted LLMs from Table~\ref{tab:subtask3-fallacywise-ablation}, analyzed in §\ref{appendix:st3:fallacy-classification}, and primarily stems from a high recall for detecting this fallacy.
For all other fallacy classes, GPT~4 achieves a substantially higher recall, leading to an overall higher performance in terms of F1-score, given the mostly similar precision across fallacy classes (cf. §\ref{appendix:analysis:fallacy-wise-classification-perfortmance-multilabel}; Figure~\ref{fig:analysis:model-comparison-per-fallacy-with-context-pr}). 
We observe the biggest difference for \emph{Ambiguity} and \emph{Impossible Expectations}, which are frequently detected by GPT~4, but never by LLaMA~2. The same was observed when prompting LLMs to predict fallacy classes applied by the gold fallacious premises (cf. §\ref{appendix:st3:fallacy-classification}; Figures~\ref{fig:figure-t3-fallacy-wise-confusion_llama} \& \ref{fig:figure-t3-fallacy-wise-confusion_gpt}), suggesting that the differences were inherent to the LLMs.
Interestingly, both LLMs perform best on the most frequent fallacy classes despite no fine-tuning involved.

\subsection{Allowing Multiple Predictions}
\begin{figure}[]
\small
    \centering
    \includegraphics[width=\linewidth]{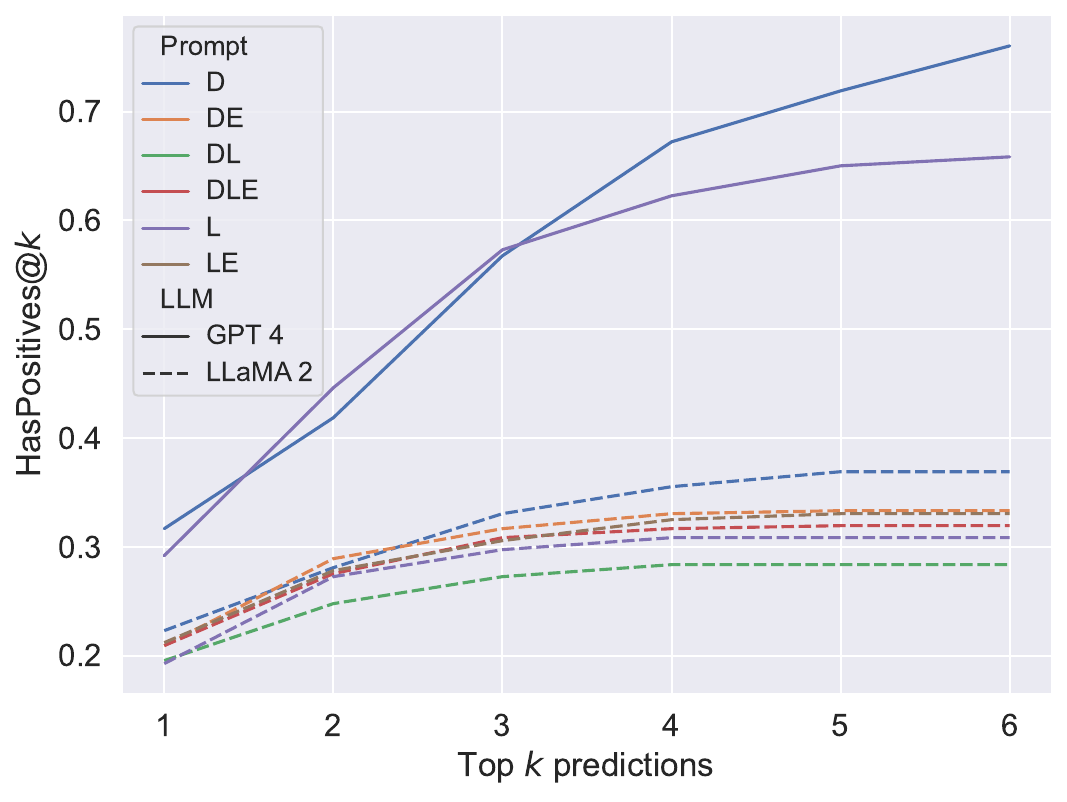}
    \caption{\textbf{Relaxed Fallacy Detection:} Performance when a fallacy is considered as correct, if the model predicts a fallacy class within the top \emph{k} results that matches any of the gold interchangeable fallacy classes.}  
    \label{fig:figure-acc-over-k}
\end{figure}

Generally, we assume that our annotators identified the most fitting fallacies that should be among the top model predictions. However,
different applicable fallacy classes may exist and our annotations cannot guarantee full recall. To address this, in Figure~\ref{fig:figure-acc-over-k}, we evaluate all LLMs in a more lenient setting. We borrow the HasPositive@$k$ metric from \citet{shaar-etal-2020-known} and consider a detected fallacy class correct, if any of the top $k$ predicted fallacy classes is accurate.
This approach avoids penalizing models for predicting different fallacies, as long as they also predict a gold fallacy class.
The results demonstrate a consistent improvement in the performance of GPT~4 as more predictions are considered. In contrast, LLaMA~2, in most cases, fails to predict the gold fallacy, even within the top 6 predictions. This confirms that GPT~4's superiority on this task is not a result of subtle selection bias for the top-ranked fallacy class, but arises from its better ability to identify the required fallacy classes. GPT~4 especially benefits from improving the detection of \emph{Fallacy of Exclusion} and \emph{False Equivalence} when increasing $k$ (cf. §\ref{appendix:detection_performance_over_k}), which account for 45.3\% of all 550 interchangeable fallacies in \dataset{}, and nearly doubles the accuracy when considering the top 3 predictions. The argument-level performance (Arg@$k$) peaks at a maximum of 89.0\% for GPT~4 and 59.7\% for LLaMA 2 for $k=6$. 

\subsection{Human Evaluation}
\label{appendix:experiments:manual-eval}
We manually evaluated 240 predictions from the main experiments in Table~\ref{tab:subtask3-contextwise-generation} (60 per LLM with (\emph{D}) and (\emph{L}) prompts; 50\% for correctly and incorrectly classified fallacies based on P@1).
%\footnote{Published alongside \dataset{}.} 
Table~\ref{tab:human-eval-main-only} shows the estimated overall results. Additional results and details are provided in §\ref{appendix:experiments:manual-eval}. Our human judgment found the fallacies produced by GPT~4, in particular (\emph{L}), the most plausible. Yet, the predicted fallacy class often did not match the premise.
Overall, we observed a major quality difference of generated premises across both LLMs, with LLaMA~2 often repeating parts of the input. NLI-S showed the strongest correlation with human judgements (Pearson $r$=0.209; p-value=$0.001$; cf. §\ref{appendix:experiments:manual-eval}). Due to the complexity of this task and its automatic evaluation, we echo \citet{schlichtkrull2023averitec} who argue that human evaluation is necessary for robustness.
\begin{table}[]
\small
    \centering
    \begin{tabular}{l| cc}
    \toprule
    \textbf{Model} & \textbf{Applicable premise} & \textbf{Correct class}  \\
    \midrule

    LLaMA~2 (\emph{L}) & 0.167& 0.040 \\
    LLaMA~2 (\emph{D})  &0.233 & 0.107 \\ 

    \midrule
    GPT~4 (\emph{L}) &\textbf{0.867} & \textbf{0.503} \\
    GPT~4 (D) &0.674 & 0.481 \\ 
    \bottomrule

    \end{tabular}
    \caption{\textbf{Human Evaluation:} Assessment if the generated fallacious premises are \emph{applicable} to bridge the reasoning gap, and if the predicted fallacy \emph{class} is applied by the generated and applicable premise.}
    \label{tab:human-eval-main-only}
\end{table}

\section{Discussion}

Following suggestions in \citet{schlichtkrull-etal-2023-intended}, we outline how our research contributes to combating misinformation. The analyzed \emph{data subjects} and \emph{data actors} are social media users. For responsible applications, we emphasize that \emph{data owners} should ideally have domain expertise, recognizing that any system will inevitably be imperfect.
We strictly did not ask the models to assign an overall rating of the claim's veracity.  Instead, we kept the user in the loop for decision-making and only assisted by outlining the fallacious discrepancies between the cited publication and the claim. Clearly communicating the inaccuracies behind a claim is important for effective debunking \citep{schmid2019effective, lewandowsky2020debunking} and can help to increase digital literacy, which is important for building resilience against misinformation \citep{lewandowsky2021countering,musi2022developing}. While previous approaches taught digital literacy using serious games \citep{roozenbeek2020breaking,musi2023developing} that require active participation, we envision a system that supports passive consumers of social media.

\section{Conclusion and Future Work}
We introduced \dataset{}, a novel dataset to combat real-world misinformation that misrepresents scientific publications. We proposed a novel task formulation to automatically reconstruct the fallacious reasoning through logical arguments based on the cited publication's content. 
We showcased \dataset{} as a testbed for evaluating the reasoning abilities of LLMs. Our experiments on two 
LLMs demonstrated the 
potential for reconstructing fallacious arguments. 
In future work, we plan to use \dataset{} with different LLMs, domains, and languages.

\section*{Limitations}
To reconstruct fallacious arguments, we solely relied on the expertise of a single fact-checking organization. \dataset{} is limited to this organization's selected claims, topics, and biases. While fallacies are derived from reasoning flaws detected by the HFC, separating fallacious reasoning from valid reasoning is not always clear-cut. Generalizing or abstracting from specific observations is an essential part of reasoning \citep{bennett2012logically} and some argue that fallacy theory in general has limited applicability for real-world claims \citep{boudry2015fake}.
When selecting claims for fact-checking, the virality of claims is a major factor \citep{arnold2020challenges}. It is, therefore, likely that information about the claim, and why it is inaccurate may have been acquired by the LLMs during pretraining \citep{magar-schwartz-2022-data}, similarly to leaked evidence effects observed in fact-checking \citep{glockner-etal-2022-missing}. While \dataset{} addresses real-world misinformation, it is just one step toward detecting such fallacies:  Our design choices exclude joint use of multiple publications to derive a claim. Assessing a claim by its cited source as done in this work is necessary but insufficient; verifying a claim in the real world requires consultation with complementary sources and domain experts \citep{silverman2014verification}. Our approach requires knowledge of the misrepresented publication, which may not always be provided together with the claim.\footnote{Although, by not providing evidence for the claim, the \emph{Evading the Burden of Proof} fallacy applies.}
Moreover, \dataset{} does not consider the original content of the misrepresented publication, but relies on paraphrasing from HFC articles, which is not available in real-world applications. Finally, \dataset{} comprises pure misinformation, and our results offer no insight into model performance over accurate claims. For practical utility, fallacy detection systems must discern whether a fallacy is present before selecting the specific type of fallacy. We note that including unbiased accurate claims is challenging, as they likely differ in topic, specificity, and may cite multiple scientific publications.  Due to these limitations, neither the tested models nor any derived from \dataset{} in this form should be directly applied in the real world.

 \section*{Ethics Statement}
The objective of this work, to combat misinformation and to increase the public resilience to it, is ethically uncritical and beneficial to society. Nevertheless, our work bears the danger of undesired side effects. Although our task definition is clearly bound to the content used when deriving a claim, our evaluation may favor models that align with the best knowledge available during COVID-19, which makes up the majority of our dataset. Yet, scientific knowledge may change over time, which will not be  reflected in \dataset{}. Moreover, we task the models to produce fallacious reasoning. This is important to explain the fallacious reasoning behind a claim for debunking \citep{lewandowsky2020debunking}, yet it may also be misused by malicious actors. 
Nevertheless, we argue that our work is rather a further demonstration of how generating fallacies in a controlled setup can be used for good, and aligns with previous work that generated misinformation to improve NLP-based approaches \citep{zellers2019defending, huang-etal-2023-faking, alhindi2023large}.
We did not take any steps to anonymise the collected data. All claims in \dataset{} are taken from HFC articles which often focus on claims by public figures. We neither contacted the individuals making the claim, nor the HFC. Following \citet{schlichtkrull2023averitec} we will remove claims from \dataset{} upon request by any individual that stated the claim, is subject of the claim or created the HFC article.

\section*{Acknowledgments}
% Adding ATHENE
 This research work has been funded by the German Federal Ministry of Education and Research and the Hessian Ministry of Higher Education, Research, Science and the Arts within their joint support of the National Research Center for Applied Cybersecurity ATHENE,
 % SQUARE
and by the the German Research Foundation (DFG) as part of the UKP-SQuARE project (grant GU 798/29-1). 
% YH  funding from TU
Yufang Hou is supported by the Visiting Female Professor Programme from TU Darmstadt.
% MG Grant fpr GPT 
We gratefully acknowledge the support of Microsoft with a grant for access to OpenAI GPT models via the Azure cloud (Accelerate Foundation Model Academic Research).
We are grateful to our dedicated annotators who helped to create \dataset{}.
Finally, we wish to thank Jan Buchmann, Nils Dycke, Aniket Pramanick, Luke Bates and Jing Yang for their valuable feedback on an early draft
of this work. 
\newpage

%\bibliography{literature,custom}
\bibliography{mybib}
\bibliographystyle{acl_natbib}
\appendix
\label{sec:appendix}

\newpage

\section{Dataset Construction I: Selecting Misrepresented Scientific Publications From HFC Articles}

\subsection{URL Filtering}
\label{appendix:dataset:step1:filtering}
For reproducibility we collect all HTML webpages via the Wayback Machine\footnote{\url{https://archive.org/web/}}.
We apply the following filtering on URLs that may be relevant to our instances.
We initially only include URLs with the following top level domains: 
\begin{itemize}
    \item \emph{``.gov'', ``.org'', ``.int'', ``.edu'', ``.gov.uk'', ``.org.uk'', ``.gov.au'', ``.org.nz'', ``.edu.au'', ``.gov.in'', ``.org.au'', ``.ac.uk''}.
\end{itemize}
We remove commonly occurring fact-checking organizations that are within the applied filtering: \emph{``fullfact.org'', ``www.poynter.org'', ``factcheck.org'', ``npr.org''}.
We finally add known publishers of scientific content that would otherwise be removed via our filtering step:
\begin{itemize}
    \item \emph{``nature.com'', ``jamanetwork.com'', ``thelancet.com'', ``researchgate.net'', ``academic.oup.com'', ``bmj.com'', ``onlinelibrary.wiley.com'',``www.mdpi.com'', ``www.ijidonline.com'', ``link.springer.com'', ``sciencedirect.com'', ``tandfonline.com'', ``journals.lww.com'', ``cell.com'', ``papers.ssrn.com'', ``cebm.net'', ``thejournal.ie'', ``cebm.ox.ac.uk'', ``elsevier.com'', ``biomedcentral.com'', ``journalofinfection.com'', ``journals.sagepub.com'', ``scientificamerican.com'', ``pfizer.com'', ``www.the-scientist.com'', ``www.cancer.net'', ``www.ema.europa.eu''}
\end{itemize}
Finally, we keep archived URLs 
\begin{itemize}
    \item \emph{``archive.is'', ``archive.ph'', ``archive.md'', ``archive.vn'', ``perma.cc'', ``archive.fo''}
\end{itemize}
as it is unknown from the surface form if it refers to a scientific publication or not.

\subsection{Annotation Process}
\label{appendix:dataset:step1:process}
\begin{figure}
\small
    \centering
    \includegraphics[width=\linewidth]{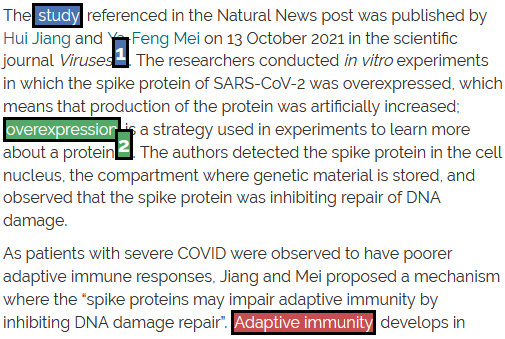}
    \caption{
    \textbf{Annotation (Step 1):}
    A preprocessed HFC article. Links for annotation are highlighted in color.}  
    \label{fig:figure-t1-highlight}
\end{figure}

 We selected 8,695 links for annotation, using a curated list of reputable scientific publishers and top-level domains (cf. §\ref{appendix:dataset:step1:filtering}).
 Few, if any, links in a fact-checking article point to a misrepresent scientific publication. HFC articles must be assessed in their entirety as critical statements may be scattered. Further, articles may cover multiple related claims, or various subclaims, each possibly misrepresenting different publications. For efficient annotation, we grouped up to 8 distinct links per fact-checking article into one HIT, highlighting each link in a different color (Figure~\ref{fig:figure-t1-highlight}). This resulted in 1,385 HITs for annotation.

Each highlighted link was given to the annotators within the original context of the HFC article, to decide if the link points to a scientific publication that is misrepresented by a non-true claim as discussed in the article at hand. For each misrepresented publication, annotators had to provide triplets consisting of
(1) a non-true claim that misrepresents
(2) a scientific publication, and
(3) a justification of the flawed reasoning.
 Each triplet requires an extracted statement from the article as justification, explaining how the document could be misused to support the claim ($S \Rightarrow \overline{c}$) and why it does not ($S \not\Rightarrow \overline{c}$). 
 Consecutive HITs present the same HFC article to prevent context switching. Annotators label a link as ``misrepresented'' only if the claim explicitly relied on the linked publication's content. Indirect misrepresentation (e.g., through a press release) is also classified as ``misrepresented'' as the claim relied on the same content. Remaining links are grouped into three sub-categories during annotation (cf. §\ref{appendix:dataset:step1:annotation}), but these are collapsed to ``not applicable'' during dataset compilation and discarded from further annotation steps. 

 In a pilot study with all 927 selected links from 50 fact-checking articles we found that  single annotation instead of double annotation (Krippendorff's~$\alpha$ of 0.728; cf. §\ref{appendix:dataset:step1:pilot}) results in a minor loss in recall only. Hence, each instance is annotated by a single annotator. Note that redundant annotations mainly impact the recall of misrepresented scientific documents. Data quality remains unaffected, as errors in link selection are rectified in later annotation tasks.  In total, we identified 208 scientific publications labeled as ``misrepresented'' across 150 HFC articles. 107 articles contained only one misrepresented publication, while 43  articles contained 2-4 misrepresented publications.

\paragraph{Definitions.}
 When selecting the triplets consisting of
(1) a non-true claim that misrepresents
(2) a scientific publication, and
(3) a justification of the flawed reasoning we consider the following definitions:
We consider a claim as \emph{non-true} and \emph{misrepresenting} if it is not fully accurate (e.g., including \emph{partly true} or \emph{misleading} claims) and lack a valid entailment relationship ($\not \Rightarrow$) with the cited document. 
We exclude documents from the misrepresented category if the claim can be validly inferred from an incorrect source or if refutation requires additional external evidence. We consider a publication as \emph{scientific} if it is published in a scientific venue, may be submitted to such a venue (preprints), or constitutes a scientific report from a credible institution, (e.g., annual CDC reports) and include non-peer-reviewed documents because they can be misrepresented before being accepted.

\subsection{Link Annotation: Fine-grained labels}
\label{appendix:dataset:step1:annotation}
Annotators categorize each link into one of four categories:
\begin{itemize}[noitemsep]
    \item \textbf{Misrepresented:} This category is designated for scientific publications that are explicitly misrepresented by a non-true claim. Such publications may be referenced either directly or indirectly, for example, through a press release.
    \item \textbf{Misrepresentable:} This category is assigned to publications that were not misrepresented but  have the potential to be misrepresented. This occurs when the HFC discuss related scientific documents for a comprehensive overview. While these documents could be susceptible to misrepresentation by similar claims, they haven't been misrepresented by the claimant.
    \item \textbf{Maybe Misrepresentable:} Annotators can choose this category when uncertain. Uncertainty may stem from ambiguity regarding a document's scientific status or doubts about misrepresentation.
    \item \textbf{Not Applicable:} This category applies to all other links not covered by the previous categories.
\end{itemize}
Annotators must provide an explanation for labels other than ``not applicable''. In \dataset{}, we focus exclusively on real-world misinformation involving genuinely misrepresented scientific documents. Instances not labeled as ``misrepresented'' are collapsed into the ``not applicable'' class and excluded. While we exclude links labeled ``misrepresentable'' or ``maybe misrepresentable'' from \dataset{}, we provide all annotations using the fine-grained taxonomy.

\subsection{Link Annotation: Pilot Study and Final Results}
\label{appendix:dataset:step1:pilot}
\paragraph{Agreement.}
The annotators, alongside one author, annotated all 221 links from 16 randomly chosen fact-checking articles. The inter-annotator agreement, assessed with Krippendorff's $\alpha$, was 0.360. Disagreement primarily arose from cases initially marked as ``misrepresentable'' or ``maybe misrepresentable'' later grouped into the ``not applicable'' category  (as per §\ref{appendix:dataset:step1:annotation}). 
Using the grouped labels, we calculated binary inter-annotator agreement between ``misrepresented'' labels and ``not applicable'' labels. This resulted in an inter-annotator agreement of 0.751. The annotators then double-annotated all 706 links from additional 34 randomly selected fact-checking articles, achieving comparable inter-annotator agreement of 0.728.

\paragraph{Single annotations are sufficient.}
We assess the value of having two annotations versus one using the double-annotated data. Specifically, if at least one annotator labels an instance as ``misrepresented'' we classify it as such. A single annotator identifies 78.3\% of the same instances as ``misrepresented''. When we also consider instances labeled as ``misrepresentable'' by the single annotator, 95.5\% of the presumed ``misprepresented'' double-annotated instances are detected. These additional cases were labeled as ``misrepresentable'' only by the second annotator, indicating more uncertainty. To reduce the workload while maintaining sufficient coverage, all remaining instances were annotated by a single annotator.

\paragraph{Results.}
In total, we found 208 (2.4\%) scientific publications labelled as ``misrepresented'', 425 (4.9\%) labelled as ``misrepresentable'', and 596 (6.9\%) labelled as ``maybe misrepresentable''. The remaining 7,466 (85.9\%) links were unrelated to our problem.

\section{Dataset Construction II: Fallacious Argument Reconstruction}

\subsection{Fallacious Reasoning for Misrepresented Science}
\label{appendix:dataset:fallacies}
\begin{table*}[]
\footnotesize{
    \centering
    \begin{tabularx}{\textwidth}{XX}
    \toprule
    \textbf{Definition} & \textbf{Logical Form} \\
    \toprule
    \multicolumn{2}{l}{\textbf{\textsc{Ambiguity}}} \\
    When an unclear phrase with multiple definitions is used within the argument; therefore, does not support the conclusion. & \textit{Claim X is made. Y is concluded based on an ambiguous understanding of X.}\\
    \midrule
    
    \multicolumn{2}{l}{\textbf{\textsc{Equivocation}} (merged with \textbf{\textsc{Ambiguity}})} \\
    When the same word (here used also for phrase) is used with two different meanings. Equivocation is a subset of the ambiguity fallacy. & \textit{Term X is used to mean Y in the premise. Term X is used to mean Z in the conclusion.} \\
    \midrule

    \multicolumn{2}{l}{\textbf{\textsc{Impossible Expectations}} / \textbf{\textsc{Nirvana Fallacy}}} \\
    Comparing a realistic solution with an idealized one, and discounting or even dismissing the realistic solution as a result of comparing to a “perfect world” or impossible standard, ignoring the fact that improvements are often good enough reason. & \textit{X is what we have. Y is the perfect situation. Therefore, X is not good enough.} \\
    \midrule

    \multicolumn{2}{l}{\textbf{\textsc{False Equivalence}}} \\
    Assumes that two subjects that share a single trait are equivalent. & \textit{X and Y both share characteristic A. Therefore, X and Y are [behave] equal. }\\
    \midrule

    \multicolumn{2}{l}{\textbf{\textsc{False Dilemma}}} \\
    Presents only two alternatives, while there may be another alternative, another way of framing the situation, or both options may be simultaneously viable. & \textit{Either X or Y is true.}  \\
    \midrule
    
    \multicolumn{2}{l}{\textbf{\textsc{Biased Sample Fallacy}}} \\
Drawing a conclusion about a population based on a sample that is biased, or chosen in order to make it appear the population on average is different than it actually is. & \textit{Sample S, which is biased, is taken from population P. Conclusion C is drawn about population P based on S.} \\
\midrule
    
    \multicolumn{2}{l}{\textbf{\textsc{Hasty Generalization}}}\\
    Drawing a conclusion based on a small sample size, rather than looking at statistics that are much more in line with the typical or average situation. &
    \textit{Sample S is taken from population P. Sample S is a very small part of population P. Conclusion C is drawn from sample S and applied to population P.}   \\
    \midrule

    \multicolumn{2}{l}{\textbf{\textsc{False Cause Fallacy}} (use as \textbf{\textsc{Causal Simplification})}} \\    
    Post hoc ergo propter hoc — after this therefore because of this. Automatically attributes causality to a sequence or conjunction of events. & \textit{A is regularly associated with B; therefore, A causes B.}  \\
    \midrule

\multicolumn{2}{l}{\textbf{\textsc{Single Cause Fallacy}} (use as \textbf{\textsc{Causal Simplification)}}} \\
Assumes there is a single, simple cause of an outcome. & \textit{X is a contributing factor to Y. X and Y are present. Therefore, to remove Y, remove X.} \\
\midrule

    \multicolumn{2}{l}{\textbf{\textsc{Fallacy of Composition}}} \\
    Inferring that something is true of the whole from the fact that it is true of some part of the whole. & \textit{A is part of B. A has property X. Therefore, B has property X.} \\
    \midrule

    \multicolumn{2}{l}{\textbf{\textsc{Fallacy of Division}} (merged with \textbf{\textsc{Fallacy of Composition})}} \\
    Inferring that something is true of one or more of the parts from the fact that it is true of the whole. & \textit{A is part of B. B has property X. Therefore, A has property X.} \\
    \midrule
    
    \multicolumn{2}{l}{\textbf{\textsc{Fallacy of Exclusion}} / \textbf{\textsc{Cherry Picking}} / \textbf{\textsc{Slothful Induction}}} \\
When only select evidence is presented in order to persuade the audience to accept a position, and evidence that would go against the position is withheld (Cherry Picking). Ignores relevant and significant evidence when inferring to a conclusion (Slothful Induction -- focus on neglect). & \textit{Evidence A and evidence B is available. Evidence A supports the claim of person 1. Evidence B supports the counterclaim of person 2. Therefore, person 1 presents only evidence A.}  \\

     \bottomrule
     
    \end{tabularx}
    }
    \caption{Fallacy Overview. Definition and logical form taken from \citet{bennett2012logically} and \citet{cook2018deconstructing}.}
    \label{tab:fallacy-overview}
\end{table*} 
\begin{table*}[]
    \centering
    \footnotesize{
    \begin{tabularx}{\textwidth}{X}
    \toprule
    \textbf{\textsc{Ambiguity}} \\
    \textit{It is said that we have a good understanding of our universe.  Therefore, we know exactly how it began and exactly when.}
    \\
    \midrule
    
    \textbf{\textsc{Equivocation}}\\
\textit{A feather is light. What is light cannot be dark. Therefore, a feather cannot be dark.} \\
    \midrule

    \textbf{\textsc{Impossible Expectations}} / \textbf{\textsc{Nirvana Fallacy}} \\
\textit{Seat belts are a bad idea. People are still going to die in car crashes.} \\
    \midrule

    \textbf{\textsc{False Equivalence}}\\
 \textit{They are both Felidae, mammals in the order Carnivora, therefore there's little difference between having a pet cat and a pet jaguar.}\\
    \midrule

    \textbf{\textsc{False Dilemma}} \\
    \textit{I thought you were a good person, but you weren’t at church today.}  \\
    \midrule
    
    \textbf{\textsc{Biased Sample Fallacy}} \\
    \textit{Based on a survey of 1000 American homeowners, 99\% of those surveyed have two or more automobiles worth on average \$100,000 each.  Therefore, Americans are very wealthy.} \\
\midrule
    
    \textbf{\textsc{Hasty Generalization}}\\
    \textit{My father smoked four packs of cigarettes a day since age fourteen and lived until age sixty-nine.  Therefore, smoking really can’t be that bad for you.}   \\
    \midrule

    \textbf{\textsc{False Cause Fallacy}}  \\    
\textit{Every time I go to sleep, the sun goes down.  Therefore, my going to sleep causes the sun to set. }  \\
    \midrule

\textbf{\textsc{Single Cause Fallacy}}\\
\textit{Smoking has been empirically proven to cause lung cancer. Therefore, if we eradicate smoking, we will eradicate lung cancer. } \\
\midrule

    \textbf{\textsc{Fallacy of Composition}} \\
\textit{Hydrogen is not wet.  Oxygen is not wet.  Therefore, water (H2O) is not wet.} \\
    \midrule

    \textbf{\textsc{Fallacy of Division}}  \\
\textit{His house is about half the size of most houses in the neighborhood. Therefore, his doors must all be about 3 1/2 feet high.} \\
    \midrule
    \textbf{\textsc{Fallacy of Exclusion}} / 
    \textbf{\textsc{Cherry Picking}} /  \textbf{\textsc{Slothful Induction}} \\
 \textit{Employer: ``It says here on your resume that you are a hard worker, you pay attention to detail, and you don’t mind working long hours.''}\\ \textit{Andy: ``Yes sir.''}\\ \textit{Employer: ``I spoke to your previous employer.  He says that you constantly change things that should not be changed, you could care less about other people’s privacy, and you had the lowest score in customer relations.''}\\
 \textit{Andy: ``Yes, that is all true, as well.''}  \\

     \bottomrule
     
    \end{tabularx}}
    \caption{Fallacy Examples (taken from \citet{bennett2012logically}).}
    \label{tab:fallacy-overview-examples}
\end{table*} 

\paragraph{Fallacy Inventory Selection.}
To select a suitable fallacy inventory, we begin by examining the fallacies employed by \citet{cook2018deconstructing} as they pertain to misinformation within the scientific domain. A distinction lies in the relation to science, as they focus on climate-change denial whereas our focus lies on the misrepresentation of scientific documents that seemingly \emph{support} the claims.  Consequently, we exclude fallacies like \emph{Red Herring} which divert attention from opposing arguments, or \emph{Fake experts}, which contradicts our requirement for credible evidence. We select the remaining fallacies by examining instances of misrepresentation of scientific publications, guided by the collection of logical fallacies from \citet{bennett2012logically}. An overview of all selected fallacies can be found in Table~\ref{tab:fallacy-overview}, along with examples in Table~\ref{tab:fallacy-overview-examples}. 
\paragraph{Merged Fallacies.}
After annotation, we merge several fallacy classes due to difficulties in differentiation based solely on the information from the fact-checking article, or because they share similar traits and one is very infrequent:
\begin{itemize}[noitemsep]
    \item \textbf{Fallacy of Division/Composition}: Combines \emph{Fallacy of Division} and \emph{Fallacy of Composition} as both involve generalizations through the \emph{part-of} relationship.
    \item \textbf{Causal Oversimplification}: Merges  \emph{Single Cause} and \emph{False Cause}as they are often indistinguishable based solely on the fact-checking article.
    \item \textbf{Ambiguity}: Combines \emph{Ambiguity} with its subtype \emph{Equivocation}, which relies on the same vocabulary in the claim and premises, a detail not accessible during annotation.
\end{itemize}
Fallacies annotated as \emph{Other} were resolved into one of the existing fallacy classes. This was always possible, as \emph{Fallacy of Exclusion} can be applied in almost all cases (as it ignores compromising content (or \emph{context}) of $S$.

\paragraph{Fallacy Inventory Discussion.}
Several fallacies, such as \emph{False Causality}, \emph{Hasty Generalization}, \emph{False Dilemma} or \emph{Ambiguity} are present in most existing NLP fallacy inventories in some form. A distinction of our selected fallacies lies in our exclusive focus on corrupted support ($\not\Rightarrow$) relationships. For this reason, we exclude several fallacies that are commonly used in fallacy detection datasets and important for propaganda techniques \citep{da-san-martino-etal-2019-fine,piskorski-etal-2023-multilingual} or misinformation in general \citep{musi2022developing,alhindi-etal-2022-multitask}:
\begin{enumerate}[noitemsep]
    \item Fallacies that \textbf{attack} (e.g. \emph{Ad Hominem}).
    \item Fallacies that \textbf{divert} (e.g. \emph{Strawman Fallacy}).
    \item Fallacies that use \textbf{manipulation techniques} like slogans or emotional language (e.g. \emph{Appeal to Emotion}).
    \item Fallacies that utilize \textbf{non-credible evidence} (e.g. \emph{False Authority}), or \textbf{omit evidence} altogether (e.g. \emph{Evading the Burden of Proof}).
\end{enumerate}

\paragraph{Fallacy Inventory Comparison.}
In contrast to other works \citep{musi2022developing,alhindi-etal-2022-multitask}, a strong focus lies on a detailed analysis of generalization fallacies. We employ various specific generalization fallacies, such as \emph{Hasty Generalization}, \emph{Biased Sample Fallacy} and \emph{False Equivalence}.
The fallacy of \emph{Cherry Picking} has been addressed in prior research. However, our interpretation of this fallacy aligns more closely with \emph{Slothful Induction}. This distinction arises because our focus is not on presenting selectively chosen information but rather on the omission of crucial aspects of the study that weaken the claim's validity.
We find \emph{False Analogy} as employed by e.g. \citet{alhindi-etal-2022-multitask}, to be comparable to \emph{False Equivalence}. Both fallacies can be applied to the same problems. Without access to the claimant's specific reasoning, we find it impossible to prioritize one over the other when generating fallacious premises. 
For similar reasons, we adopt a broad definition of the \emph{False Dilemma} fallacy, which assumes that only two options (or outcomes) exist when, in reality, more options are available. We consolidate it with the fallacy of \emph{Affirming the Disjunct}, which assumes an ``either/or'' possibility among different options, even when these options are not mutually exclusive. Despite their differences (in \emph{False Dilemma} the options are indeed mutually exclusive but not exhaustive), they share similar characteristics in exhibiting black-and-white thinking and may only differ in their specificity of the explicit reasoning that is unavailable to us.

\subsection{Argument Reconstruction Guidelines}
\label{appendix:dataset:argument-construction}
Annotators should base their reconstruction of fallacious arguments on content in the fact-checking article. The final argument should be:
\begin{itemize}[noitemsep]
    \item \textbf{Comprehensive:}  All fallacies identified by the fact-checker should be incorporated.
    \item \textbf{Self-contained:} Subsequent steps should not necessitate the use of the fact-checking articles.
\end{itemize}
For all text generation tasks our annotators utilize Grammarly\footnote{\url{https://www.grammarly.com}}, integrated into the Surge AI tool, to ensure high-quality text.
Each of the previously detected 208 misrepresented links was provided to the annotators together with the the preprocessed fact-checking article with highlighted links, and the justification for labeling the link as ``misrepresented'' (from §\ref{appendix:dataset:step1:process}) and annotators were tasked to reconstruct all parts of the fallacious argument.
Annotations were conducted in batches of 30-40 arguments per annotator per week with weekly meetings, adhering to agile annotation principles \citep{alex-etal-2010-agile}. Each argument was independently assessed by at least two annotators. 
At the end of annotation, annotators reviewed their own annotations, excluding the last batch, to rectify errors resulting from initial guideline misunderstandings and enhance consistency.

\paragraph{Claim Rewriting.}
\begin{figure}[h]
\small
    \centering
    \includegraphics[width=\linewidth]{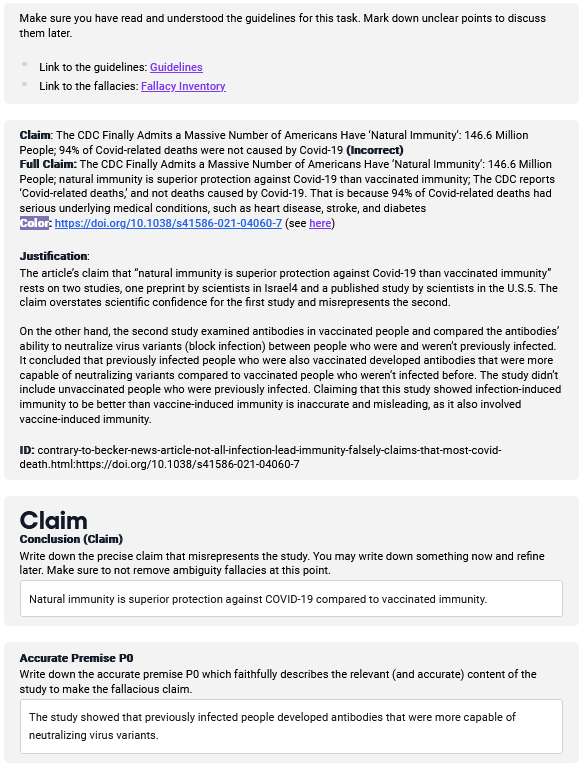}
    \caption{Claim annotation interface from Surge AI.}  
    \label{fig:figure-t2-claim}
\end{figure}

Fact-checking articles might discuss multiple related claims or complex arguments with subclaims. Annotators should first understand the main claim of the fact-checking article and the misrepresented scientific document.  The misrepresenting claim $\overline{c}$ should be formulated as such that $S \cup \overline{P} \Rightarrow \overline{c}$. Annotators should use the main claim of the fact-checking article if possible, and make minimal changes if necessary. While the formulation of \(\overline{c}\) is a prerequisite for detecting fallacious reasoning lines $R$, the validity of \(S \cup \overline{P} \Rightarrow \overline{c}\) can only be checked after constructing the argument. Therefore, annotators should re-evaluate \(\overline{c}\) after identifying all fallacies. The annotation interface, including a link to the pre-processed fact-checking article and relevant information is shown in Figure~\ref{fig:figure-t2-claim}.

\paragraph{Accurate Premise Writing.}
The accurate premise $p_0$ provides a correct description of the misrepresented scientific document $S$.  Its purpose is to offer logical support for the claim ($p_0 \Rightarrow \overline{c}$) but it falls short due to the presence of fallacious reasoning.
Fact-checkers always include an accurate description of the misrepresented scientific document. Annotators must locate all relevant information and formulate $p_0$ s.t.
\begin{enumerate}[noitemsep]
    \item The wording is as precise as possible and uses the HFC vocabulary.
    \item All accurate content that strengthens  $p_0 \Rightarrow \overline{c}$ is included.
    \item Any accurate content that weakens $p_0 \Rightarrow \overline{c}$ is excluded. 
\end{enumerate}
We guide annotators to include information in the accurate premise $p_0$ only if it \emph{increases} the plausibility of $\overline{c}$. Any information (from $S$) that \emph{decreases} plausibility is paraphrased as $s_i$ and is part of $R_i$. 
The publication context $s_i$ is optional and required only if additional information beyond the given $p_0$ is necessary.

\paragraph{Hidden Premise Writing.}
Fact-checkers explain how the claim $\overline{c}$ relates to the scientific publication \(S\). This may include additional knowledge not found in $\overline{c}$ or \(S\).\footnote{This relies on subjective judgment since annotators aren't required to read the scientific document.} 
For example, to understand why the claim ``Cucumber kills lung cancer cells.'' was made based on the scientific finding that ``cucubitacin B promoted lung tumor cell death.'' one must know that cucumbers contain cucubitacin B. This information is likely not provided by the misrepresented publication itself.
Annotators can provide any number of hidden premises which are concise, accurate statements that complement \(S\) and are essential in understanding the connection between \(\overline{c}\) and \(S\). Each hidden premise should be a single sentence derived from the fact-checking article.

\paragraph{Fallacy Class Selection Preference}
\label{appendix:dataset:fallacies-selection-priorities}
Annotators are directed to prioritize fallacies that engage with the content (all fallacies except \emph{Fallacy of Exclusion}) rather than ignoring crucial aspects.
We allow interchangeable fallacies with distinct classes, but we instruct annotators to prioritize more specific fallacies over broader ones. When multiple fallacies share the same flawed reasoning, annotators should select the most specific fallacy class. For example, when a conclusion is drawn from a biased sample, it can be labeled as the \emph{Biased Sample Fallacy}. Alternatively, it might be seen as the \emph{Single Cause Fallacy} assuming that the properties for which the sample is biased do not impact the conclusion. In this example, both fallacy classes do not differ in their applied reasoning, but only in their level of specificity. Therefore, the more specific \emph{Biased Sample Fallacy} should be preferred.\footnote{This is different to the example in Figure~\ref{fig:figure-alternative-fallacies-example}, in which different fallacies apply different reasoning.}. We provided a taxonomy to the annotators to specify how to choose the more specific fallacy class if multiple apply.

\paragraph{Fallacious Premise Writing.}
\begin{figure}
\small
    \centering
    \includegraphics[width=\linewidth]{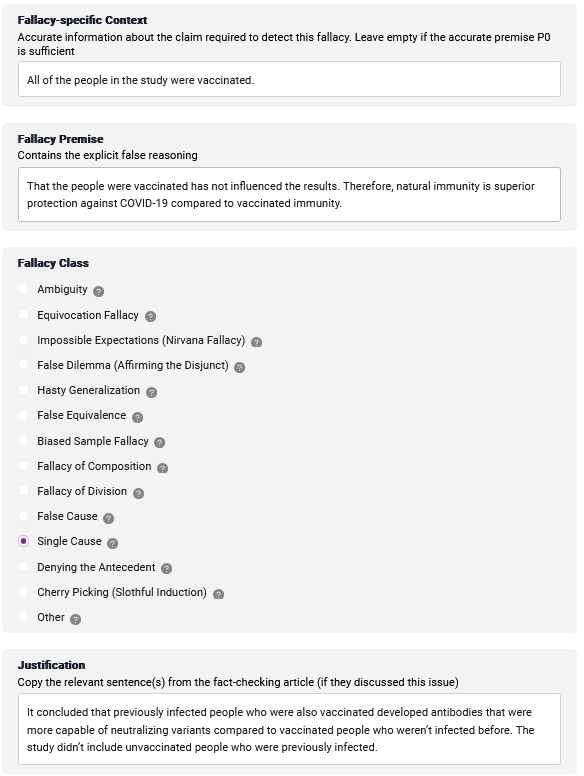}
    \caption{Fallacy annotation interface from Surge AI.}  
    \label{fig:figure-t2-fallacies}
\end{figure}

Annotators should thoroughly review the fact-checking article to identify all sections discussing the claim \(\overline{c}\) misrepresenting the scientific publication \(S\). These discussions may be distributed throughout the article. Annotators must focus solely on fallacies discernible from the content of \(S\). This excludes other scientific documents used by the HFC to refute \(\overline{c}\) or assessing the credibility of \(S\) (e.g., if  $S$ is retracted or inaccurate). Each fallacy should be separately formulated by the annotator and must be accompanied by a justification extracted from the fact-checking article.
The context for each fallacy must be constructed akin to \(p_0\), emphasizing the weakening of the claim rather than its strengthening. The fallacious premise must align with the selected fallacy class and make the fallacious reasoning explicit.
Given that selecting the fallacious reasoning is subjective since different fallacious premises \(p_i\) can re-instantiate \(S \cup p_i \Rightarrow \overline{c}\), annotators are permitted to formulate multiple alternative variants that they consider plausible. The annotation interface for a fallacy is visualized in Figure~\ref{fig:figure-t2-fallacies}.
After constructing all fallacies, annotators are tasked with reviewing their own constructed argument to ensure the coherence of the fallacies, claim, and accurate premise, in their HIT.

\subsection{Argument Consolidation}
\label{appendix:dataset:consolidation}
Our primary annotator, with the most project experience, handled the consolidation process. The consolidator aligned all annotated fallacious reasoning lines, and selected the best verbalized candidate for each $\overline{c}$, $\overline{p}_i$ and $s_i$, or paraphrased multiple such candidates into an improved version.
Interchangeable fallacies with different classes were preserved. Only clearly unjustified fallacy annotations were discarded or corrected if possible. Each consolidated argument underwent double-checking by an author. The consolidator and one of the authors collaborated on a final round of fallacious premise curation. Due to different URLs used to link to the same scientific document, some arguments with duplicate annotations were merged using normalized URLs (cf. §\ref{appendix:dataset:document-annotation}), resulting in 193 distinct fallacious arguments $\mathcal{A}$. Individual annotators might discard arguments during annotation if they cannot reconstruct a fallacious argument, for example because of insufficient information provided by the HFC or violations of the credibility assumption of the cited publication (we analyze the impact of such cases where fewer annotations are available in §\ref{sec:fallacy-coverage}). In seven cases no annotator could reconstruct $\mathcal{A}$, and two more fallacious arguments were discarded during consolidation.

\subsection{Scientific Document Annotation.}
\label{appendix:dataset:document-annotation}
\begin{figure}[h]
\small
    \centering
    \includegraphics[width=\linewidth]{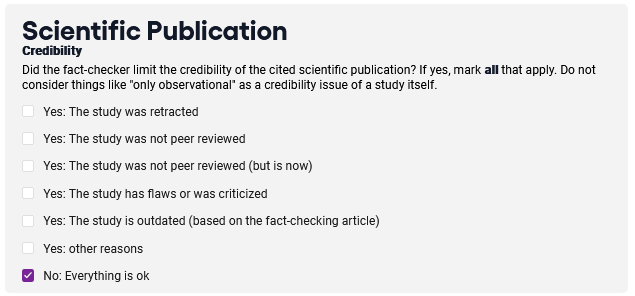}
    \caption{Credibility annotation via Surge AI.}  
    \label{fig:figure-t2-credibility}
\end{figure}

Fact-checkers may use varying links of the same publication within their fact-checking articles. To detect duplicates these links must be normalized. To assess the content of publicly available studies, we require them to be available via PMC.
Annotators are tasked to find a URL pointing to the misrepresented scientific publication $S$ and select the first applicable URL of the following list:
\begin{enumerate}[noitemsep]
    \item PMC (available as \emph{fulltext})
    \item sciencedirect (available as \emph{fulltext})
    \item Original Publisher (available as \emph{fulltext})
    \item PubMed (available as \emph{abstract only})
    \item PMC (available as \emph{abstract only})
    \item sciencedirect (available as \emph{abstract only})
    \item Original Publisher (available as \emph{abstract only})
    \item as-is
\end{enumerate}
We also asked whether the HFC highlighted flaws in the study itself (see Figure~\ref{fig:figure-t2-credibility}). 

\section{Dataset Analysis}

\subsection{Publication Credibility}
\label{appendix:publication-credibility}
\begin{table}[]
\small
    \centering
    \begin{tabular}{l| c c}
    \toprule
    \textbf{Annotation} & \textbf{Documents} & \textbf{Krippendorff's $\alpha$} \\
    \toprule
    Flawed/Criticised & 39 & 0.504 \\
    Preprint & 28 & 0.582 \\
    Retracted & 2 & 0.665 \\
    Outdated & 1 & 0.0 \\
    \midrule
    \textbf{Any from Above} & \textbf{61} & \multirow{ 2}{*}{\textbf{0.577}} \\
    \textbf{Credible} & \textbf{123} &  \\
    
     \bottomrule
     
    \end{tabular}
    \caption{\textbf{Publication Credibility:} Results of the credibility annotations of the misrepresented publications per fallacious argument alongside with the inter-annotator agreement per label. 
    }
    \label{tab:credibility_results}
\end{table} 
\citet{augenstein-2021-determining} identifies two key challenges in science communication: (1) assessing the credibility of scientific publications, and (2) preventing the misrepresentation of a study's findings. This study addresses the second challenge, while assuming the credibility. To examine to which degree our assumption holds, we report the credibility annotations of the used publications in Table~\ref{tab:credibility_results}. Note that a publication may exhibit more than one credibility issue (e.g., a preprint, that was criticised) 
We consider a publication $S$ as ``credible'' only if no annotator detected any aspect compromising its credibility in the HFC article, providing a conservative estimate.
For 123 arguments, no credibility violations were identified. Even identified violations do not necessarily mean that the publication is not credible or not misrepresented by $\overline{c}$. For instance, preprints lack formal community approval but may still be credible. At the time of annotation, 16 out of 28 publications marked as preprints by the HFC were accepted. Further, criticism of a study does not make it scientifically unsound. In fact, some critics highlight omitted limitations that could lead to misunderstandings similar to the misrepresentations studied in this work.
The overall agreement in distinguishing credible documents from those with any credibility issue is 0.577 (Krippendorff's $\alpha$). 
\begin{figure}[h]
\small
    \centering
    \includegraphics[width=\linewidth]{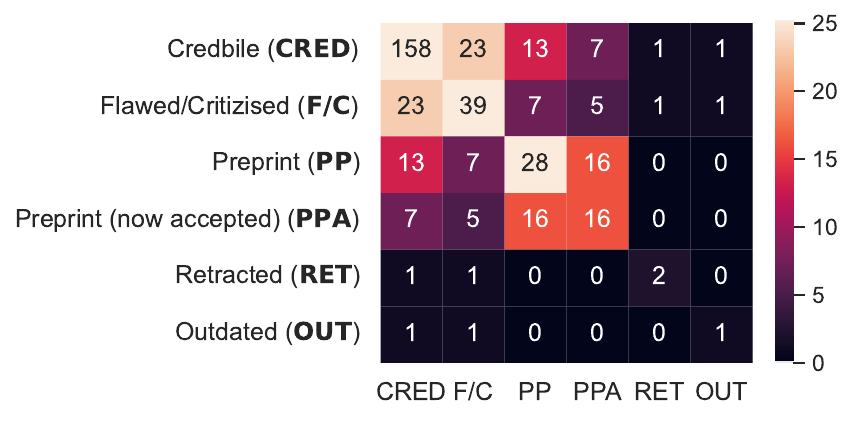}
    \caption{Heatmap of co-occurring credibility annotations per misrepresented document a fallacious argument.}  
    \label{fig:confusion-matrix-credibility}
\end{figure}

Figure~\ref{fig:confusion-matrix-credibility} displays the 
co-occurring credibility annotations assigned to fallacious arguments. We count an occurrence of a credibility label if at least one annotator assigned it to the misrepresented document, and, hence, ignore duplicate annotations of the same label.

\subsection{Claims about the COVID-19 infodemic}
\label{appendix: claim-topics}
\begin{figure}
\small
    \centering
    \includegraphics[width=\linewidth]{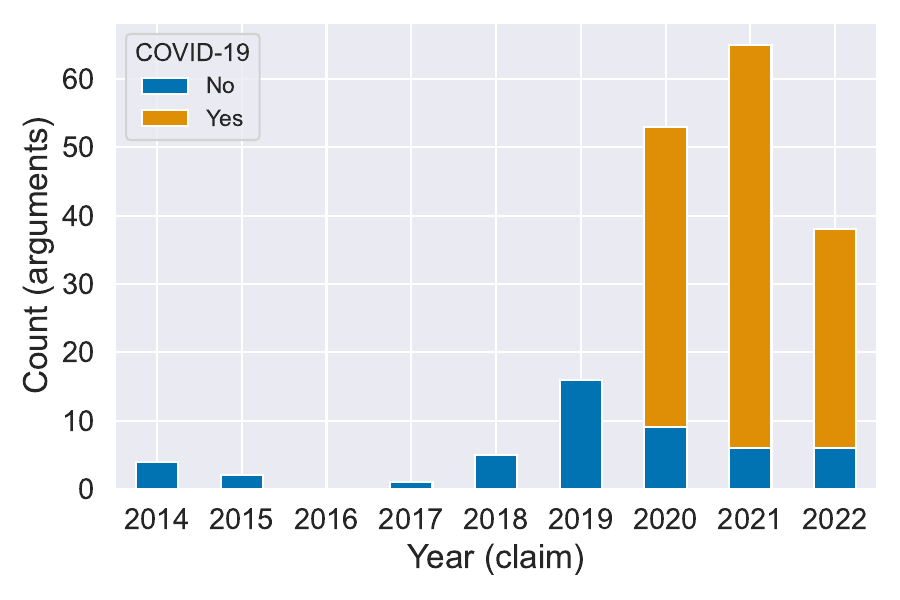}
    \caption{\textbf{COVID-19 Arguments:} Collected arguments per year and whether they constitute COVID-19 related misinformation.}  
    \label{fig:figure-arg-over-years}
\end{figure}

\dataset{} comprises fallacious arguments from 2014 to 2022. We manually evaluate each argument for its association with COVID-19 misinformation, considering context provided by the HFC, rather than explicit COVID-19 mentions. For instance, we consider the claim ``Ivermectin sterilizes most men who take it'' as COVID-19-related because it only exists due to the misinformation about Ivermectin as a COVID-19 cure. Figure~\ref{fig:figure-arg-over-years} illustrates argument distribution over time and their COVID-19 relevance.

\subsection{On the Coverage of Fallacies}
\label{sec:fallacy-coverage}
\begin{figure}[h]
\small
    \centering
    \includegraphics[width=\linewidth]{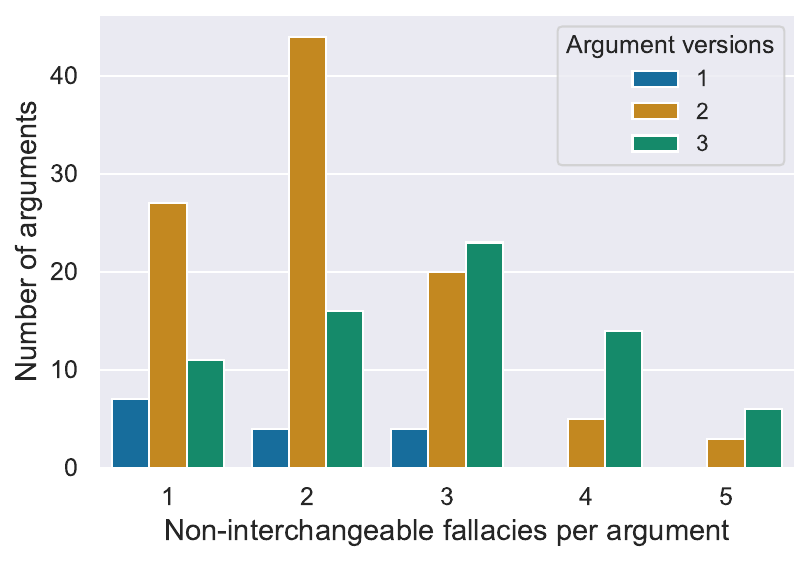}
    \caption{\textbf{Argument Annotations:} Number of successfully constructed arguments (\emph{y-axis}) compared to the number of assigned annotators who manually reconstruct each argument (\emph{color})  per distinct fallacious reasoning lines $R_i$ in the argument post consolidation (\emph{x-axis}). The number of arguments shows if one (\emph{blue}), two (\emph{orange}) or three (\emph{green}) distinct arguments versions of the same argument were constructed by the annotators.}   
    \label{fig:figure-t2-number-problems-per-arg}
\end{figure}

In some cases, annotators may decide that an argument cannot be reconstructed and discard the HIT. This can for example happen, if the annotator considers the information provided by the HFC insufficient to reconstruct the argument, or if the cited publication itself is non-credible. 
 Table~\ref{tab:problem-coverage} compares the annotation assignments and successful argument reconstructions. A total of 153 arguments were reconstructed by all assigned annotators. In 25 cases, one annotator, and in 6 cases, two annotators could not reconstruct the argument. Given the limited number of annotations and ambiguity when assigning fallacy classes, we aim to shed light into the recall of our detected fallacies. To this end, we compare the number of successfully reconstructed fallacious arguments with the number of detected fallacies (excluding interchangeable fallacies) in Figure~\ref{fig:figure-t2-number-problems-per-arg}.  More annotators generally lead to more identified fallacies. The majority of arguments with successful reconstructions of two and three annotators comprise two or three distinct fallacies respectively.

The analysis indicates that \dataset{} may lack comprehensiveness of fallacies, in arguments with fewer annotations. However, this does not affect our main objective since annotators are likely to identify the most severe fallacies, highlighted by the HFC, that violate \emph{strong inductive support}  (cf. §\ref{sec:preliminaries:inductive}). A single study, under specific conditions, rarely gives unconditional support for any claim $c$ without potential generalizations. While not necessarily or severely fallacious, these generalizations may be less emphasized by HFC, which makes them harder to detect as fallacious reasoning for the annotators.
In real-world texts, identifying different fallacies among annotators is not unusual \citep{da-san-martino-etal-2019-fine}, and distinguishing between fallacious and accurate reasoning can be subtle \citep{boudry2015fake}.
\begin{table}[]
\small
    \centering
    \begin{tabular}{cc | c }
    \toprule
    \multicolumn{2}{c}{\emph{Annotators}} & \emph{Result}\\
    \textbf{Assigned} & \textbf{Successful} & \textbf{Num. Arguments}  \\
    \toprule
    2 & 1 & 9 \\
    3 & 1 & 6 \\
    3 & 2 & 16 \\
    \multicolumn{2}{r|}{\textbf{Incl. failed reconstructions}} & \textbf{31} \\
    \midrule
    2 & 2 & 83 \\
    3 & 3 & 70 \\
    \multicolumn{2}{r|}{\textbf{No failed reconstruction}} & \textbf{153} \\

     \bottomrule
     
    \end{tabular}
    \caption{\textbf{Annotation Assignments:} Number of successfully reconstructed argument versions by the number of initially assigned annotators.}
    \label{tab:problem-coverage}
\end{table} 

\section{Experiments}
\subsection{Fallacious Premise Evaluation}
\label{appendix:experiments:metrics}

We use BERTScore \citep{bert-score} F1 with version 0.3.13, based on DeBERTa \citep{he2020deberta}, fine-tuned on MNLI \citep{williams2018mnli}, as recommended at the time of this writing.\footnote{\url{https://github.com/Tiiiger/bert_score}} For the NLI-based metric, we use the predicted probability of the entailment label, following \citep{honovich-etal-2022-true}. We use their provided T5-XXL model \citep{raffel2020exploring}, fine-tuned on SNLI \citep{bowman-etal-2015-large}, MNLI \citep{williams2018mnli}, FEVER \citep{thorne-etal-2018-fever}, Scitail \citep{khot2018scitail}, PAWS \citep{zhang-etal-2019-paws}, and VitaminC \citep{schuster-etal-2021-get}. The predicted probability for the token ``\texttt{1}'' (representing \emph{entailment}) is used for evaluation.

Multiple interchangeable reference fallacious premises $\overline{p}_{i,j}$ may exist. Therefore, before evaluating the generated fallacious premise $\hat{\overline{p}}_{i,1}$, it needs to be matched with a reference text. For the first-ranked predicted fallacy by the model, which consists of a fallacious premise and fallacy class ($\hat{\overline{p}}_{i,1}$, $\hat{f}_{i,1}$), we use the gold fallacious premise $\overline{p}_{i,j}$ as the reference text if the accompanied gold fallacy class $f_{i,j}$ matches the predicted fallacy class $\hat{f}_{i,1}$.
 In the absence of matches based on the fallacy class, we select the $\overline{p}_{i,j}$ with the highest cosine similarity to $\hat{\overline{p}}_{i,1}$ from all interchangeable fallacious premises $\overline{p}_{i,j}$. Cosine similarity is measured using the sentence embeddings produced by SBERT~\citep{reimers-2019-sentence-bert} (\texttt{all-mpnet-base-v2}).

\subsection{Prompt Engineering for Fallacy Generation and Classification}
\label{appendix:st3:argument-reconstructuion}

    \begin{table}[]
\small
    \centering
    \begin{tabular}{l|cc }
    \toprule
    \textbf{Template} & \textbf{P@1} & \textbf{Arg-1} \\
    \midrule
    p1-basic (D) & 0.208 &  0.467 \\
p2-support (D) & 0.222 &  0.467 \\
p3-undermine (D) & 0.208 &  0.467 \\
p4-connect (D) & \textbf{0.278} &  \textbf{0.533} \\
p5-auto (D) & 0.194 &  0.433 \\
p5-auto-connect (D) & 0.264 &  0.467 \\
\midrule
p1-basic (DE) & 0.208 &  0.467 \\
p2-support (DE) & 0.194 &  0.433 \\
p3-undermine (DE) & 0.208 &  0.467 \\
p4-connect (DE) & \textbf{0.292} &  \textbf{0.600} \\
p5-auto (DE) & 0.194 &  0.433 \\
p5-auto-connect (DE) & 0.278 &  0.500 \\
\midrule
p1-basic (DL) & 0.208 &  0.467 \\
p2-support (DL) & 0.222 &  0.500 \\
p3-undermine (DL) & 0.236 &  0.533 \\
p4-connect (DL) & \textbf{0.361} &  \textbf{0.700} \\
p5-auto (DL) & 0.250 &  0.567 \\
p5-auto-connect (DL) & 0.306 &  0.633 \\
\midrule
p1-basic (DLE) & 0.236 &  0.500 \\
p2-support (DLE) & 0.236 &  0.533 \\
p3-undermine (DLE) & 0.250 &  0.533 \\
p4-connect (DLE) & \textbf{0.306} &  \textbf{0.600} \\
p5-auto (DLE) & 0.222 &  0.500 \\
p5-auto-connect (DLE) & 0.236 &  0.500 \\
\midrule
p1-basic (L) & 0.181 &  0.367 \\
p2-support (L) & 0.208 &  0.433 \\
p3-undermine (L) & 0.236 &  0.500 \\
p4-connect (L) & \textbf{0.278} &  \textbf{0.567} \\
p5-auto (L) & 0.194 &  0.400 \\
p5-auto-connect (L) & 0.236 &  0.433 \\
\midrule
p1-basic (LE) & 0.208 &  0.467 \\
p2-support (LE) & 0.250 &  0.567 \\
p3-undermine (LE) & 0.222 &  0.500 \\
p4-connect (LE) & 0.236 &  0.500 \\
p5-auto (LE) & 0.222 &  0.500 \\
p5-auto-connect (LE) & \textbf{0.264} &  \textbf{0.600} \\
    \bottomrule

    \end{tabular}
    \caption{\textbf{Argument Reconstruction:} Prompt tuning using LLaMA~2 (70B)}
    \label{tab:subtask3-fallacywise-generation:dev}
\end{table}

The only existing prompts related to fallacy detection stem from \citet{alhindi-etal-2022-multitask}. Yet, they were (a) not used in zero-shot experiments, instead relying on fine-tuning, and (b) were used to only classify the fallacy within a single text. Since our task differs and demands a zero-shot setup, we manually select novel prompts for the task. 
We initially experiment on few instances to derive promising prompt templates. We assess the performance of each prompt with different combinations of (\emph{D}, \emph{L}, and \emph{E}) using LLaMA~2 on the 30 arguments of the validation split in Table~\ref{tab:subtask3-fallacywise-generation:dev}.
All prompts and hyper-parameters are listed in §\ref{appendix:reprocudibility}.

\subsection{Premise Evaluation on Correct/Incorrect Fallacy Classes}
\label{appendix:experiments:premise-eval-correct-incorrect}

\begin{table*}[]
\small
    \centering
    \begin{tabular}{l|cccc | cccc }
    \toprule
    & \multicolumn{4}{c|}{\emph{Correct Fallacy Class}} & \multicolumn{4}{c}{\emph{Incorrect Fallacy Class}} \\
   \textbf{Model} & \textbf{METEOR} & \textbf{BERTScore} & \textbf{NLI-A} & \textbf{NLI-S} & \textbf{METEOR} & \textbf{BERTScore} & \textbf{NLI-A} & \textbf{NLI-S}\\
    \midrule

    LLaMA~2 (\emph{D}) & 0.214 & 0.616 & 0.181 & 0.231  & 0.224 & 0.617 & 0.106 & 0.124 \\
    LLaMA~2 (\emph{DE}) & 0.238 & 0.615 & 0.144 & 0.174 & 0.227 & 0.622 & 0.119 & 0.142  \\
    LLaMA~2 (\emph{DL}) & 0.218 & 0.608 & 0.105 & 0.120 & 0.199 & 0.618 & 0.136 & \textbf{0.148}  \\
    LLaMA~2 (\emph{DLE}) & 0.236 & 0.624 & 0.130 & 0.154 & 0.199 & 0.614 & \textbf{0.128} & 0.143 \\
    LLaMA~2 (\emph{L}) & 0.243 & 0.622 & \textbf{0.239} & \textbf{0.266} & \textbf{0.256} & \textbf{0.628} & 0.116 & 0.141  \\
    LLaMA~2 (\emph{LE}) & 0.194 & 0.605 & 0.149 & 0.168 & 0.177 & 0.610 & 0.114 & 0.125  \\
    \midrule
    GPT 4 (\emph{D}) & 0.259 & \textbf{0.636} & 0.094 & 0.200 & 0.229 & 0.611 & 0.057 & 0.092  \\
    GPT 4 (\emph{L}) & \textbf{0.264} & \textbf{0.637} & 0.109 & \textbf{0.267} & 0.227 & 0.604 & 0.046 & 0.088  \\

    \bottomrule

    \end{tabular}
    \caption{\textbf{Fallacious Premise Evaluation based on Fallacy Class Correctness:} Automatic metrics separately evaluated over fallacious premises when the predicted fallacy class was correct or not.}
    \label{tab:subtask3-contextwise-generation-correct-incorrect}
\end{table*} 
We closely examine the predictions in §\ref{sec:experiments:main-results} and analyze the generated fallacious premises under two conditions: (a) when accompanied by a correct fallacy class, and (b) when not accompanied by an accurate fallacy class. The results are presented in Table~\ref{tab:subtask3-contextwise-generation-correct-incorrect}. 
Specifically, for the metrics METEOR, BERTScore, and NLI-S, the fallacious premises produced by GPT~4 outperform those generated by LLaMA~2 when the fallacy class was correctly predicted. This differs strongly from the results when not separating correct from incorrect fallacy predictions. LLaMA~2~(\emph{D}), the previously superior model in fallacious premise generation (from the main results in Table~\ref{tab:subtask3-contextwise-generation}), maintains a good performance in this evaluation. In almost all cases, the performance of both LLMs is higher for correctly classified fallacies compared to incorrectly classified premises. This is likely attributed to the fact that we only have matching fallacious premises as reference text when the fallacy class is accurately predicted, but also indicates a certain consistency between the fallacy class and fallacious premise generation.
While we believe it is important to report the performance as done here, we chose to report the overall performance in the main results in §\ref{sec:experiments:main-results}. This decision is made because (a) models may (and do, see §\ref{appendix:experiments:manual-eval}) produce correct premises but assign the inaccurate fallacy class, and (b) to ensure comparability among models, as all scores are based on identical instances and not influenced by the model's specific fallacy classification performance.

\subsection{Class-wise analysis of the fallacy classification}
\label{appendix:st3:fallacy-classification}
\begin{figure*}
\small
    \centering
    \includegraphics[width=\textwidth]{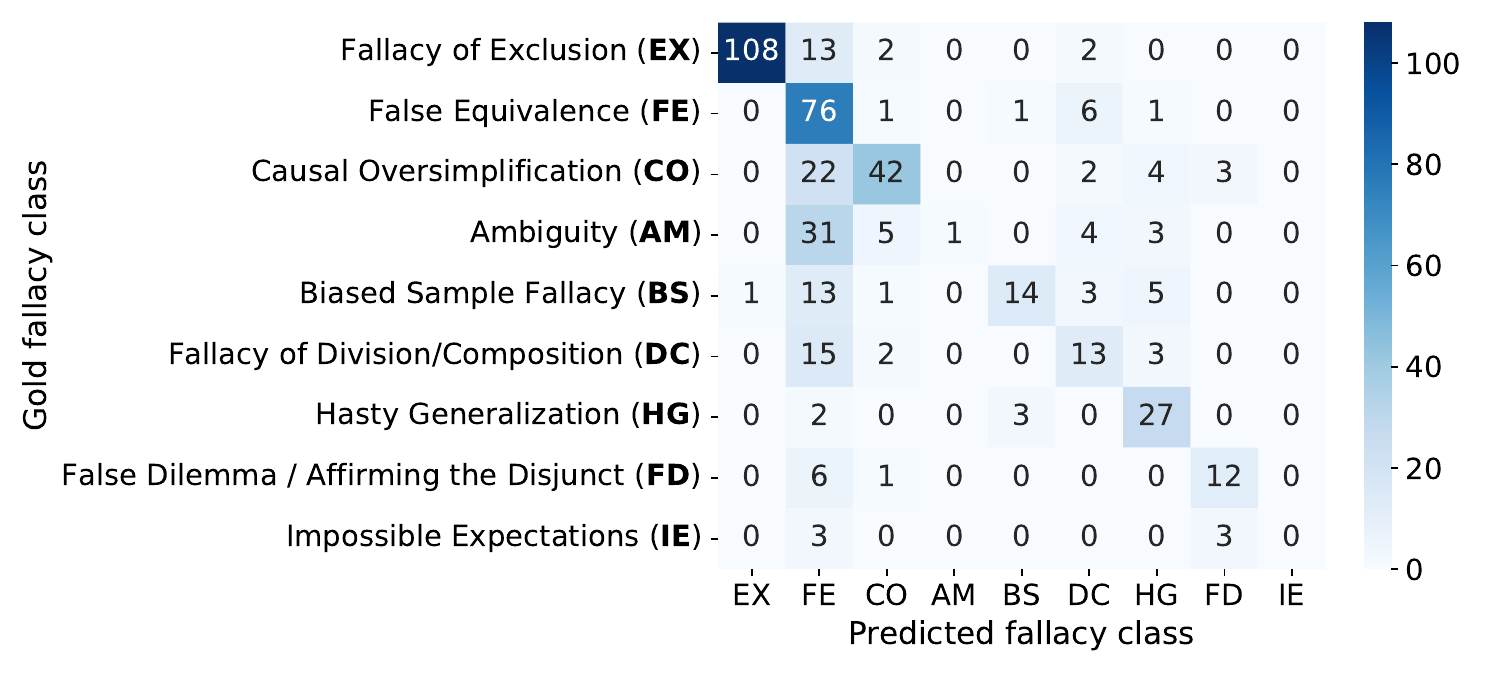}
    \caption{\textbf{Fallacy Classification for LLaMA~2} Confusion matrix based on LLaMA~2 (\emph{Logical + Example}) of predicted and gold fallacy classes when provided with $p_0$, $s_i$, $\overline{p_i}$ and $\overline{c}$.}  
    \label{fig:figure-t3-fallacy-wise-confusion_llama}
\end{figure*}

\begin{figure*}
\small
    \centering
    \includegraphics[width=\textwidth]{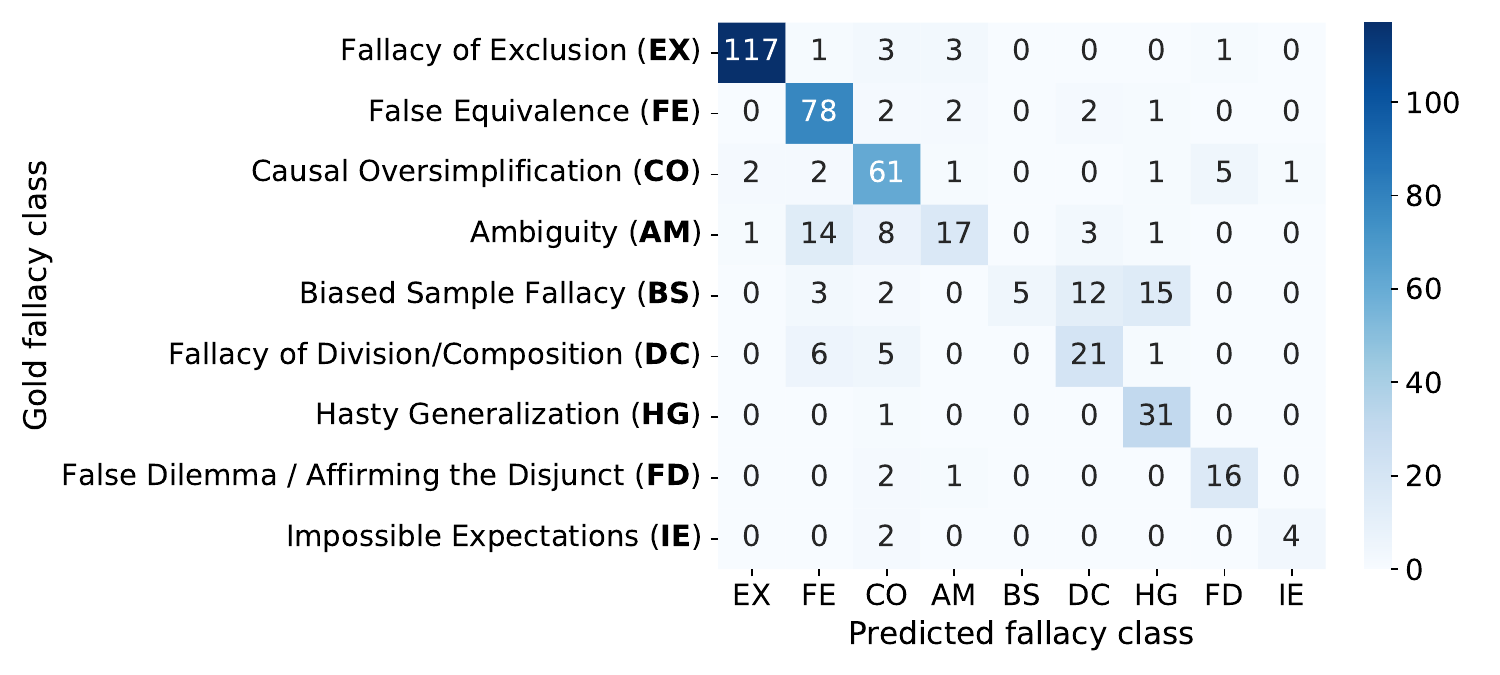}
    \caption{\textbf{Fallacy Classification for GPT~4} Confusion matrix based on GPT~4 (\emph{Logical + Example}) of predicted and gold fallacy classes when provided with $p_0$, $s_i$, $\overline{p_i}$ and $\overline{c}$.}   
    \label{fig:figure-t3-fallacy-wise-confusion_gpt}
\end{figure*}

We examine the class-wise predictions of the best-performing models concerning the fallacies applied in the gold fallacious premises $\overline{p}_i$. The confusion matrices of LLaMA~2 (\emph{LE}) and GPT~4 (\emph{LE}) are presented in Figures~\ref{fig:figure-t3-fallacy-wise-confusion_llama}~\&~\ref{fig:figure-t3-fallacy-wise-confusion_gpt}. Both LLMs exhibit strong performances across the majority of fallacy classes.
For GPT~4, the most confusion arises between \emph{Ambiguity} and \emph{False Equivalence}, as well as between \emph{Biased Sample Fallacy} and \emph{Hasty Generalization}. The commonalities between the former two fallacies align with the frequently co-occurring fallacies identified by our annotators in §\ref{sec:interchangable-fallacy-analysis}. Conversely, \emph{Hasty Generalization}, although clear-cut in our annotations, is often used interchangeably with its superclass \emph{Faulty Generalization} in the wild. We hypothesize that GPT~4 may have acquired knowledge during pretraining from discussions where claims generalizing from biased subsets are labeled as \emph{Hasty Generalization}.
Similarly, for LLaMA~2, most confusion primarily arises between \emph{False Equivalence} and \emph{Ambiguity}. However, we observe that LLaMA~2 tends to overpredict \emph{False Equivalence} in general.

\section{Experiment Analysis}
\subsection{GPT~4 (D) Fallacy Detection Performance over k}
\label{appendix:detection_performance_over_k}
\begin{figure}
\small
    \centering
    \includegraphics[width=\linewidth]{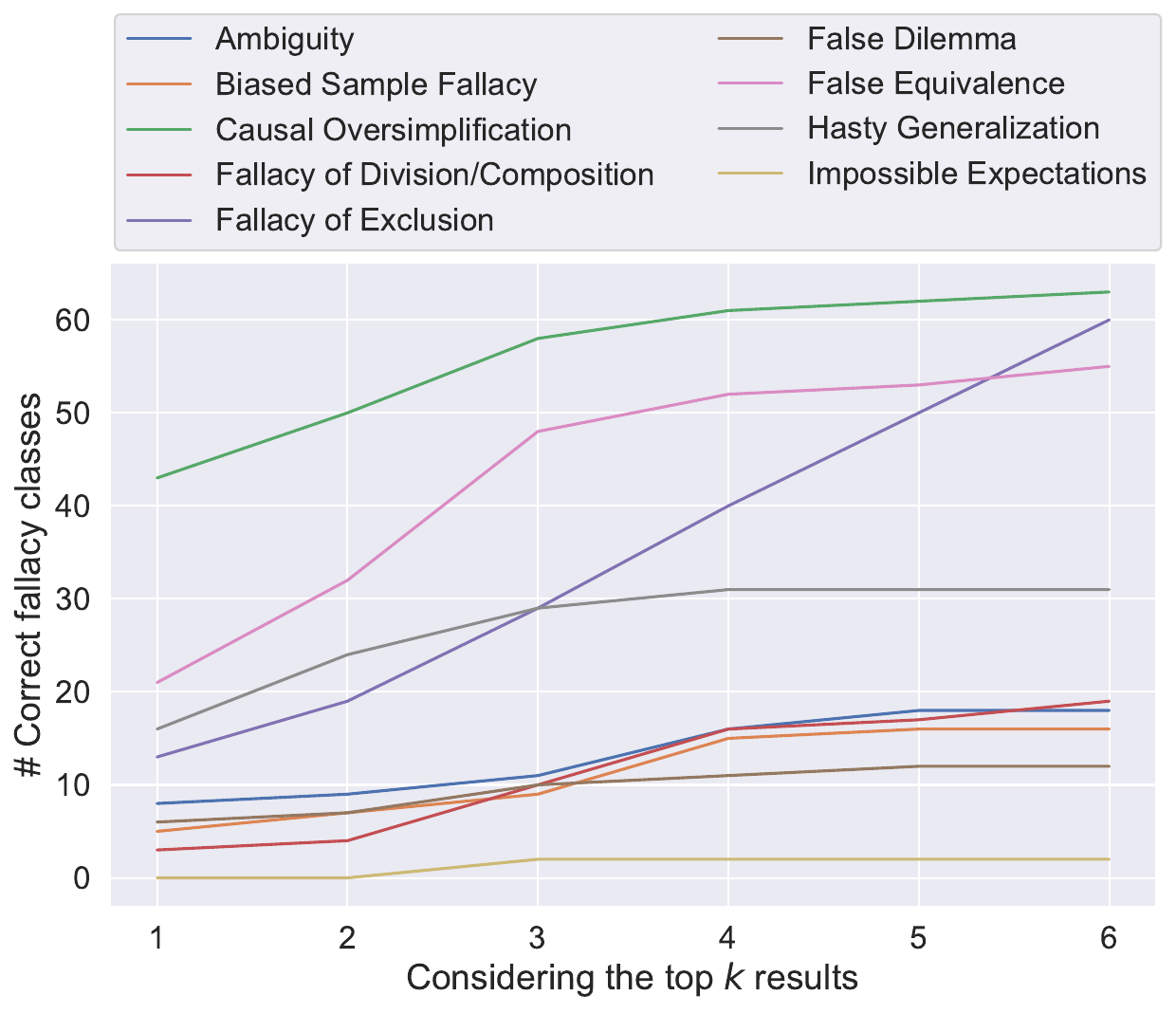}
    \caption{
    Number of correctly predicted fallacy classes of the GPT~4~(\emph{D}) model when considering all correctly predicted fallacies among the top $k$ predictions.}  
    \label{fig:gpt_4d_fallacywise_per_k}
\end{figure}

We report the number of correctly detected fallacies per distinct fallacy class in Figure~\ref{fig:gpt_4d_fallacywise_per_k} for the GPT-4 (\emph{D}) model with the best overall performance from §\ref{sec:experiments:main-results}. By allowing the model to provide $k$ additional fallacy class predictions, the number of correctly identified fallacies increases, especially for the \emph{Fallacy of Exclusion} and \emph{False Equivalence}.

\subsection{Fallacy-wise classification performance}
\label{appendix:analysis:fallacy-wise-classification-perfortmance-multilabel}
\begin{figure}
\small
    \centering
    \includegraphics[width=\linewidth]{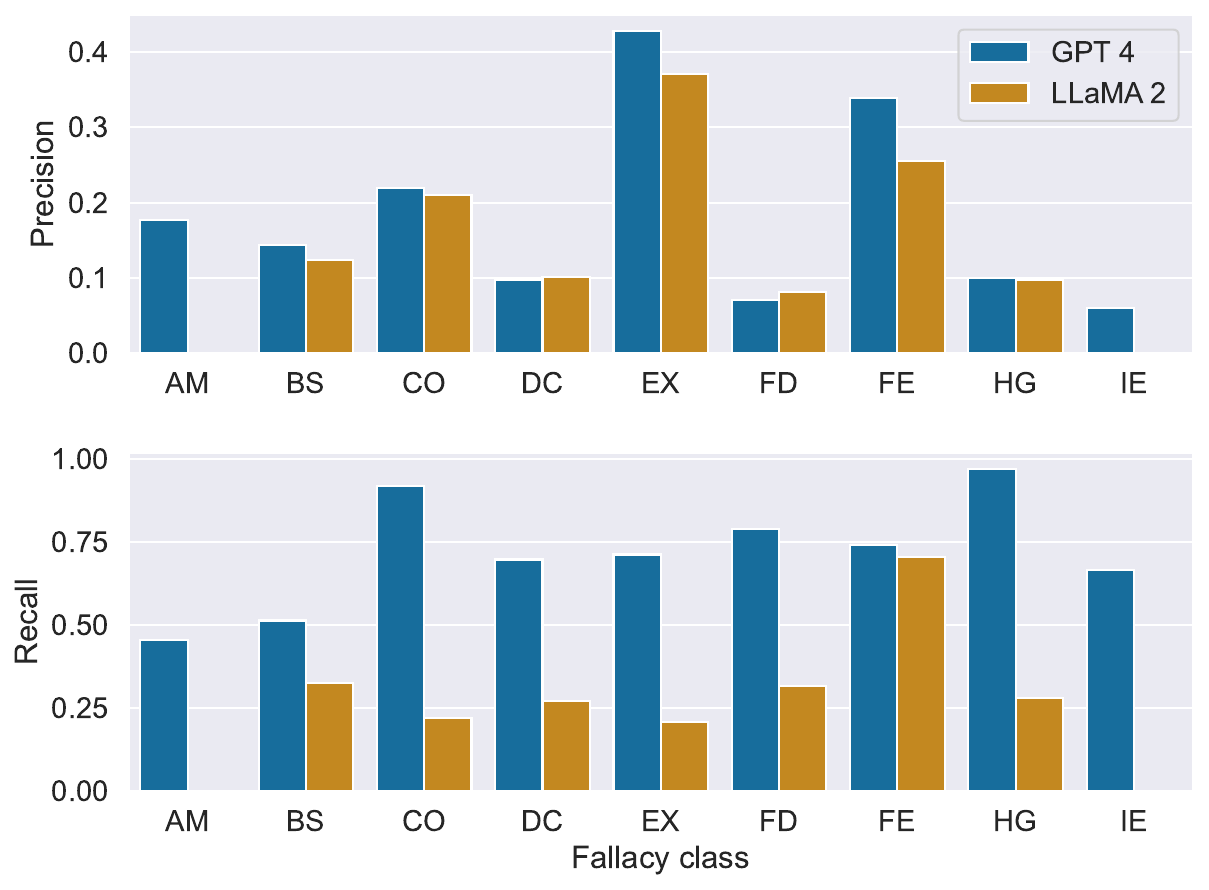}
    \caption{
    \textbf{Per-Fallacy Performance:}
    F1 score per predicted fallacy class from a multi-label multi-class classification perspective considering all predictions by the respective model.}  
    \label{fig:analysis:model-comparison-per-fallacy-with-context-pr}
\end{figure}

We visualize the precision (top) and recall (bottom) from a multi-class multi-label perspective for each distinct fallacy class of LLaMA-2 and GPT-4 (both (\emph{D})) in Figure~\ref{fig:analysis:model-comparison-per-fallacy-with-context-pr}.

\subsection{The impact of the fallacious premise}
\label{appendix:analysis:implicit-classification}
To better understand the impact of the generated fallacious premise $\overline{p}_i$, we  additionally compare and evaluate the LLMs to see how well they can predict a correct fallacy class given only the inaccurate claim $\overline{c}$, the accurate premise $p_0$ and the context $s_i$, without generating the fallacious premise $\overline{p}_i$. We compare the results with the accuracy achieved in the fallacy reconstruction (Table~\ref{tab:subtask3-contextwise-generation}), and when the fallacious premise is provided (Table~\ref{tab:subtask3-fallacywise-ablation}).
The results are depicted in Table~\ref{tab:implicit-fallacy-classification}. We report P@1 when the gold $\overline{p}_i$ is not provided to the model (i.e. for \emph{Reconstruct} and \emph{n/a}), since each of the interchangeable fallacies is equally valid. When $\overline{p}_i$ is provided (\emph{Given}), we report the stricter accuracy metric in which only the one applied fallacy is correct. Interestingly, LLaMA~2 consistently performs better when it is not required to generate the fallacious reasoning for the fallacy classification task. 
\begin{table}[]
\small
    \centering
    \begin{tabular}{l | c c c}
    \toprule
    & \multicolumn{3}{c}{\emph{Role of $\overline{p}_i$}} \\
    \textbf{Model} & \textbf{Reconstruct} & \textbf{Given} & \textbf{n/a} \\
    \midrule
    LLaMA~2 (\emph{D}) & 0.223 & 0.577 & 0.248 \\
    LLaMA~2 (\emph{DE}) & 0.209 & 0.637 & 0.264 \\
    LLaMA~2 (\emph{DL}) & 0.196 & 0.630 & 0.237 \\
    LLaMA~2 (\emph{DLE}) & 0.209 & 0.568 & 0.259 \\
    LLaMA~2 (\emph{LE}) & 0.212 & 0.645 & \textbf{0.267} \\
    LLaMA~2 (\emph{L}) & 0.193 & 0.601 & 0.262 \\
    \midrule
    GPT 4 (\emph{D}) & \textbf{0.317} & 0.738 & \textbf{0.267} \\
    GPT 4 (\emph{L}) & 0.292 & \textbf{0.744} & 0.245 \\
    \bottomrule

    \end{tabular}
    \caption{\textbf{Fallacy classification without fallacious premises:} Performance when predicting the applied fallacy class only, when required to \emph{reconstruct} the fallacious premise, when the gold fallacious premise is \emph{given} to the model, or when the fallacious premise is \emph{n/a} and does not need to be generated.}
    \label{tab:implicit-fallacy-classification}
\end{table} 
We hypothesize that this is due to the poor utility of the fallacious premises generated by LLaMA~2 (cf. manual analysis in §\ref{appendix:experiments:manual-eval}). For GPT~4, which produced substantially better fallacious premises according to our human evaluation, the model exhibits improved performance, similar to chain-of-thought prompting \citep{wei2022chain}, when tasked with generating the fallacious premise for both evaluated prompts. 

To further shed light on this, In Table~\ref{tab:gpt-per-matched-unmatched} we compare the performance of GPT~4 when (a) it was able to generate a fallacious premise that matched the reference premise according to human judgment, and (b) when it did not. We report the P@1 (or accuracy) for the GPT~4 models\footnote{We only report GPT~4 results here because we found too few instances in which LLaMA~2 generated a matching fallacious premise in §\ref{appendix:experiments:manual-eval}.} when tasked to additionally generate the fallacious premise, only classify the fallacy, or classify the fallacy given the gold $\overline{p}_i$ in Table~\ref{tab:implicit-fallacy-classification}. The results show a substantial performance difference among both evaluated splits of instances, regardless of whether the fallacious premises needed to be generated or not, suggesting that these instances are challenging for both objectives. When provided with the gold reference $\overline{p}_i$, however, the difference in performance decreases, showing that the verbalized fallacious reasoning overcomes these challenges. For both prompts, the best fallacy classification performance was achieved when the model was tasked with additionally generating the fallacious premise and did so correctly. Note that we report P@1 if the fallacious premise $\overline{p}_i$ is not provided, and accuracy otherwise.
\begin{table}[]
\small
    \centering
    \begin{tabular}{ll|cc}
    \toprule
    & & \multicolumn{2}{c}{\emph{Matching $\hat{\overline{p}}_i$}} \\
    \textbf{Model} & \textbf{Setup} & \textbf{Yes} & \textbf{No} \\
    \midrule
         &classify $f_i$ and gen. $\overline{p}_i$ & \textbf{0.880} & 0.229 \\
    GPT~4 (\emph{D}) &classify $f_i$ w/o $\overline{p}_i$ & 0.640 & 0.114  \\
         & classify $f_i$ given $\overline{p}_i$ & 0.788 & \textbf{0.689}  \\

    \midrule
         &classify $f_i$ and gen. $\overline{p}_i$ & \textbf{0.867} & 0.133 \\
             GPT~4 (\emph{L}) &classify $f_i$ w/o $\overline{p}_i$ & 0.533 & 0.133  \\
     & classify $f_i$ given $\overline{p}_i$ & 0.732 & \textbf{0.722}  \\

    \bottomrule
    \end{tabular}
    \caption{\textbf{Comparison to fallacy generation difficulty}: We separate the 60 instances from our manual analysis per GPT~4 by whether the generated fallacious premises apply the same fallacious reasoning as the reference gold fallacious premises (\emph{Yes}) or not (\emph{No}) and report the performance in three setups for both evaluated prompts.}
    \label{tab:gpt-per-matched-unmatched}
\end{table}

\subsection{Manual Evaluation of Generated Fallacies}
\label{appendix:experiments:manual-eval}
For the first ranked generated fallacious premise $\hat{\overline{p}}_{i}$ and predicted fallacy class $\hat{f}_i$ of 240 model predictions we assess if
\begin{enumerate}[noitemsep]
    \item the LLM outputs a fallacious premise (\textbf{Q1}),
    \item the generated premise $\hat{\overline{p}}_{i}$ represents an applicable premise within the argument bridges the reasoning gap based on the context provided via $s_i$ (if exists) (\textbf{Q2}),
    \item Q2 and the generated premise $\hat{\overline{p}}_{i}$ applies the predicted fallacy class $\hat{f}_i$ (\textbf{Q3}),
    \item the generated premise $\hat{\overline{p}}_{i}$ expresses the same fallacious reasoning as the reference $\overline{p}_{i}$ (\textbf{Q4}).
\end{enumerate}
We answer each question with yes/no, without awareness of the model or prompt for each prediction.
Only in a single instance, the LLM failed to generate any fallacious premise (Q1). 
We provide the results of our manual analysis (Q2 and Q3) in Table~\ref{tab:human-eval-main}. Most differences can be seen among both LLMs. Premises generated when a gold fallacy was assigned were considered valid slightly more often compared to when no gold fallacy was predicted. When the gold fallacy was predicted, the predicted fallacy class almost always matched the fallacy of the generated premise. However, when no gold fallacy class was predicted, often the predicted fallacy class did also not match the generated premise. Since we used stratified sampling when selecting the predictions, we approximate the overall performance $s$:
\[
s = \mathrm{P@1}\times h^{correct} + (1-\mathrm{P@1}) \times h^{incorrect}
\]
where $h^{correct}$ (or $h^{incorrect}$) represent the manual evaluation among all instances considered correct (or incorrect) by the automatic evaluation via P@1.
Examples are provided in Table~\ref{tab:human-eval-main-examples}. The first example contains relatively similar premises generated by the LLM and the annotators. In the second example, the LLM provides the same fallacious reasoning that applies results from mice to humans, yet the premise is much more specific and tailored to the provided content of the misrepresented publication. However, the human premise does not entail the LLM-generated premises. The third example exhibits semantically similar premises. However, the LLM-generated premise makes a causal assumption, differing from the \emph{Biased Sample Fallacy} employed in the human-written premise. In the last example, the LLM-generated premise looks similar but does not logically support the claim, which reasons that in order to stop COVID-19, we need higher temperatures (and hence climate change) since higher temperatures stop the spread of COVID-19.

We use Q4 to assess the quality of each applied metric in Table~\ref{tab:reeval-metrics}. We separately report each metric among premise pairs when the generated premises involves the same fallacious reasoning as the reference premise (score should be high), and when not (score should be low).
As commonly done \citep{banarjee2005,bert-score}, we report Pearson correlation to human judgment. 
When computing the Pearson correlation, we convert ``yes'' and ``no'' into numerical values of 1 and 0. Unlike semantic similarity, we observed that deciding whether the same reasoning flaw is verbalized is more often a binary decision than a decision on a continuous scale.

We found NLI-S to be most correlated with human judgment, while the asymmetric counterpart NLI-A exhibited the least correlation. This confirms our initial hypothesis that LLMs may produce more specific premises than our annotators (see examples in §\ref{appendix:experiments:manual-eval}, Table~\ref{tab:human-eval-main-examples}), and vice versa, which should be accounted for by the metric. 
\begin{table}[]
\small
    \centering
    \begin{tabular}{l| c c | c c}
    \toprule
    & \multicolumn{2}{c|}{\emph{Matching premises}} & \multicolumn{2}{c}{\emph{Pearson r}}\\ 
    \textbf{Metric} & \textbf{Yes} & \textbf{No} & \textbf{\emph{r}}&  \textbf{P-value}\\
    \midrule

METEOR & 0.280 & 0.227 & 0.153 & 0.017*   \\
BERTScore & 0.638 & 0.622 & 0.092  & 0.158  \\
NLI-A & 0.155 & 0.140 & 0.022  & 0.736  \\
NLI-S & 0.317 & 0.150 & 0.209  & 0.001*  \\
    
     \bottomrule
     
    \end{tabular}
    \caption{\textbf{Human evaluation (Q4):} Comparison of the used metrics reported over 68 generated premises that match the reference premise and 172 premises that do not match the reference premise. We measure the correlation with the human judgment via Person \emph{r} and mark significant results with an asterisk.    
    }
    \label{tab:reeval-metrics}
\end{table}

\begin{table*}[]
\small
    \centering
    \begin{tabular}{l| cc | cc | cc}
    \toprule
     & \multicolumn{2}{c|}{\emph{Gold Fallacy Prediction}} & \multicolumn{2}{c|}{\emph{No Gold Fallacy Prediction}}  & \multicolumn{2}{c}{\emph{Overall} (approximated)}\\
    \textbf{Model} & \textbf{Plausible} & \textbf{Matching } & \textbf{Plausible } & \textbf{Matching } & \textbf{Plausible } & \textbf{Matching } \\
    \midrule

    LLaMA~2 (\emph{L}) & 5 (17\%) &  5 (17\%) & 2 (7\%) & 1 (3\%) & \emph{0.167} & \emph{0.040} \\
    LLaMA~2 (\emph{D})  & 7 (23\%) & 7 (23\%) & 4 (13\%) & 3 (10\%) &\emph{0.233} & \emph{0.107} \\ 

    \midrule
    GPT~4 (\emph{L})  & 26 (87\%) & 26 (87\%) & 25 (83\%) & 11 (37\%)&\emph{0.867} & \emph{0.503} \\
    GPT~4 (\emph{D}) & 25 (83\%) & 24 (80\%) & 18 (60\%) & 10 (33\%)&\emph{0.674} & \emph{0.481} \\ 
    \bottomrule

    \end{tabular}
    \caption{Results of our manual analysis based on 240 ($30\times4\times2$) manually assessed model predictions. We report the number of generated premises deemed \textbf{plausible}, and the number of plausible fallacious premises in which the applied fallacy class \textbf{matches} the predicted fallacy class.}
    \label{tab:human-eval-main}
\end{table*} 

\begin{table*}[]
\small
    \centering
    \begin{tabularx}{\textwidth}{l| X|X}
    \toprule
    \textbf{Matches} & \textbf{LLM} & \textbf{Annotator} \\
    \midrule
    yes & Since THC is a compound in marijuana and HU210 is more potent than THC, it follows that marijuana has the same brain cell growth properties as HU210. & HU210 and THC have similar properties. Since HU210 increases the number of brain cells, THC also causes brain cells to grow. \\
    \midrule
    yes & Because HU210, a synthetic compound found in cannabis, increases the number of cells in the hippocampus of mice, the use of any cannabis-related substance, like marijuana, will yield the same results in humans. & Mice and humans are both mammals. Therefore, marijuana grows brain cells in humans.\\
    \midrule
    no & The fluoride in the pineal gland comes from the fluoride in toothpaste. & Elderly people are a subset of all people. Therefore, fluoride in toothpaste harms the pineal gland in all humans. \\
    \midrule
    no & Efforts to combat climate change will result in warmer average temperatures, therefore increasing the prevalence of the SARS-CoV-2 virus. & Lower SARS-CoV-2 transmission was associated with lower temperature. Therefore, lower temperature facilitates SARS-CoV-2 transmission. \\
    \bottomrule
     
    \end{tabularx}
    \caption{\textbf{Matching of Fallacious Premises:} Examples of the manual evaluation with fallacious premises produced by an LLM and by the annotators, together with our rating whether they express the same reasoning flaw or not.}
    \label{tab:human-eval-main-examples}
\end{table*}

\section{Reproducibility}
\label{appendix:reprocudibility}

\subsection{Hyperparameters}
\label{appendix: hyperparameters}
For reproducibility, experiments on LLaMA~2 use a batch size of 1, zero temperature, and a beam size of 1, with a fixed random seed set to 1. LLaMA~2 experiments were implemented using the Huggingface transformers library \citep{wolf-etal-2020-transformers}.
To address computational constraints, we employ 4-bit quantization \citep{NEURIPS2022_c3ba4962} for LLaMA~2 (70B), which we found comparable to 8-bit quantization in preliminary experiments. All prompting experiments wit LLaMA~2 models are executed on A100 GPUs. For GPT~4 we  use the GPT version=\texttt{gpt-4} and the API version=\texttt{2023-10-01-preview} with a maximum new token length of $1000$ and content filtering turned off, using the OpenAI API\footnote{\url{https://platform.openai.com/docs/api-reference}}.

\subsection{Prompts}
\label{appendix:prompts}

\subsubsection{Prompts for Fallacious Argument Reconstruction}
\label{appendix:prompts-reconstruction}
We include \emph{definitions}, \emph{logical forms} or \emph{examples} from literature \citep{bennett2012logically, cook2018deconstructing} (cf. §\ref{appendix:dataset:fallacies}). Notably, examples do not constitute real-world fallacies, but constitute educational examples that clearly outline the fallacious reasoning behind them, such as ``A feather is light. What is light cannot be dark. Therefore, a feather cannot be dark'' (cf. Table~\ref{tab:fallacy-overview-examples}).
We assess different combinations thereof in various prompt templates outlined in Figure~\ref{fig:prompt-template-argument-reconstruction}. Using this template, we evaluate various task descriptions (Figures~\ref{fig:prompt-reconstruct-basic}-\ref{fig:prompt-reconstruct-auto-connect}; key differences are highlighted in \textbf{bold}) for the different combinations of definitions, logical forms and examples.
We do not argue that these are the respective best prompts to solve the introduced task. Rather, they serve as useful baselines that  for future research that and are identical for both evaluated LLMs to allow for direct comparison as well as the analysis of the impact of \emph{definitions}, \emph{logical forms} or \emph{examples} within these prompts.

    \begin{figure*}[!ht]
    \begin{tcolorbox}[left=1pt, right=1pt, top=1pt, bottom=1pt, boxrule=0.5pt]
    \small
    \textbf{Fallacy Inventory:} \\

Ambiguity: \\
Definition 1: \emph{When an unclear phrase with multiple definitions is used within the argument; therefore, does not support the conclusion.} \\
Logical Form 1: \emph{Claim X is made. Y is concluded based on an ambiguous understanding of X.} \\
Example 1: \emph{It is said that we have a good understanding of our universe.  Therefore, we know exactly how it began and exactly when. } \\
Definition 2: \emph{When the same word (here used also for phrase) is used with two different meanings.} \\
Logical Form 2: \emph{Term X is used to mean Y in the premise. Term X is used to mean Z in the conclusion.} \\
Example 2: \emph{A feather is light. What is light cannot be dark. Therefore, a feather cannot be dark.} \\

Impossible Expectations: \\
Definition 1: \emph{Comparing a realistic solution with an idealized one, and discounting or even dismissing the realistic solution as a result of comparing to a “perfect world” or impossible standard, ignoring the fact that improvements are often good enough reason.} \\
Logical Form 1: \emph{X is what we have. Y is the perfect situation. Therefore, X is not good enough.}\\
Example 1: \emph{Seat belts are a bad idea. People are still going to die in car crashes.} \\

    \texttt{<more fallacies>} \\
    \\
    \textbf{Task:} \\
    \texttt{<task-instruction with instance>}

    \end{tcolorbox}
    \caption{\textbf{Argument Reconstruction Template:} Overall template used to reconstruct fallacious arguments in which \emph{definition}, \emph{logical form} and \emph{example} are used. We evaluate different combinations thereof, as well as different task descriptions.}
    \label{fig:prompt-template-argument-reconstruction}
    \end{figure*}

    \begin{figure*}[!ht]
    \begin{tcolorbox}[left=1pt, right=1pt, top=1pt, bottom=1pt, boxrule=0.5pt]
    \small
        Examine the following fallacious argument: \\
    Premise 1: ``\texttt{<$p_0$>}'' \\
    Premise 2: ``\texttt{<$s_i$>}'' \\
    Premise 3: ``'' \\
    Therefore: ``\texttt{<claim>}'' \\
    
    Premises 1 and 2 are sourced from the same credible scientific document.
The claim is based on the information in Premise 1.
However, Premise 2 suggests that the claim is an invalid conclusion from the scientific document. \\

Your task is to identify and verbalize the fallacious reasoning in Premise 3 (the fallacious premise) that is necessary to support the claim, despite the conflicting information in Premise 2.
Only consider fallacies from the provided fallacy inventory. \\

Present each fallacious premise along with the applied fallacy class in this format: \\

    Fallacious Premise: <fallacious premise>; Applied Fallacy Class: <applied fallacy class>. \\ 

If there are multiple applicable fallacies, list them in order of relevance.

    \end{tcolorbox}
    \caption{\textbf{P1 Basic:} Our most basic task instruction to reconstruct the fallacious argument.}
    \label{fig:prompt-reconstruct-basic}
    \end{figure*}

        \begin{figure*}[!ht]
    \begin{tcolorbox}[left=1pt, right=1pt, top=1pt, bottom=1pt, boxrule=0.5pt]
    \small
    Examine the following fallacious argument: \\
    Premise 1: ``\texttt{<$p_0$>}'' \\
    Premise 2: ``\texttt{<$s_i$>}'' \\
    Premise 3: ``'' \\
    Therefore: ``\texttt{<claim>}'' \\
    
    Premises 1 and 2 are sourced from the same credible scientific document.
The claim is based on the information in Premise 1.
However, Premise 2 suggests that the claim is an invalid conclusion from the scientific document. \\

Your task is to identify and verbalize the fallacious reasoning in Premise 3 (the fallacious premise) that is necessary to support the claim, despite the conflicting information in Premise 2.
\textbf{This reasoning should be strong enough to support the claim and counter any uncertainties raised by Premise 2.}
Only consider fallacies from the provided fallacy inventory. \\

Present each fallacious premise along with the applied fallacy class in this format: \\

    Fallacious Premise: <fallacious premise>; Applied Fallacy Class: <applied fallacy class>. \\

If there are multiple applicable fallacies, list them in order of relevance. 

    \end{tcolorbox}
    \caption{\textbf{P2 Support:} The model is tasked to produce a fallacious premise that increases the support behind the claim.}
    \label{fig:prompt-reconstruct-support}
    \end{figure*}

        \begin{figure*}[!ht]
    \begin{tcolorbox}[left=1pt, right=1pt, top=1pt, bottom=1pt, boxrule=0.5pt]
    \small
    Examine the following fallacious argument: \\
    Premise 1: ``\texttt{<$p_0$>}'' \\
    Premise 2: ``\texttt{<$s_i$>}'' \\
    Premise 3: ``'' \\
    Therefore: ``\texttt{<claim>}'' \\

Premises 1 and 2 are sourced from the same credible scientific document.
The claim is based on the information in Premise 1.
However, Premise 2 suggests that the claim is an invalid conclusion from the scientific document. \\

Your task is to identify and verbalize the fallacious reasoning in Premise 3 (the fallacious premise) that is necessary to support the claim, despite the conflicting information in Premise 2.
\textbf{This reasoning should effectively support the claim, ensuring that Premise 2 does not undermine the claim as a valid conclusion.}
Only consider fallacies from the provided fallacy inventory. \\

Present each fallacious premise along with the applied fallacy class in this format: \\

    Fallacious Premise: <fallacious premise>; Applied Fallacy Class: <applied fallacy class>. \\

If there are multiple applicable fallacies, list them in order of relevance.

    \end{tcolorbox}
    \caption{\textbf{P3 Undermine:} The task is rather phrased negatively. The model must generate the fallacious premise to avoid that $s_i$ undermines the claim.}
    \label{fig:prompt-reconstruct-undermine}
    \end{figure*}

        \begin{figure*}[!ht]
    \begin{tcolorbox}[left=1pt, right=1pt, top=1pt, bottom=1pt, boxrule=0.5pt]
    \small
Examine the following fallacious argument: \\
    Premise 1: ``\texttt{<$p_0$>}'' \\
    Premise 2: ``\texttt{<$s_i$>}'' \\
    Premise 3: ``'' \\
    Therefore: ``\texttt{<claim>}'' \\

Premises 1 and 2 are sourced from the same credible scientific document.
The claim is based on the information in Premise 1.
However, Premise 2 suggests that the claim is an invalid conclusion from the scientific document. \\

Your task is to identify and verbalize the fallacious reasoning in Premise 3 (the fallacious premise) that is necessary to support the claim, despite the conflicting information in Premise 2.
Do not repeat the claim itself, Premise 1, or Premise 2 when generating the fallacious Premise 3. \textbf{Make sure the generated Premise 3 connects Premise 1 and Premise 2 to robustly support the claim, and ensure that Premise 2 does not undermine the claim as a valid conclusion.}
Only consider fallacies from the provided fallacy inventory.\\

Present each fallacious premise along with the applied fallacy class in this format: \\

    Fallacious Premise: <fallacious premise>; Applied Fallacy Class: <applied fallacy class>. \\

If there are multiple applicable fallacies, list them in order of relevance.

    \end{tcolorbox}
    \caption{\textbf{P4 Connect:} The task definition explicitly requires the model to connect Premises 1 and 2 with the conclusion via the fallacious premise.}
    \label{fig:prompt-reconstruct-connect}
    \end{figure*}

        \begin{figure*}[!ht]
    \begin{tcolorbox}[left=1pt, right=1pt, top=1pt, bottom=1pt, boxrule=0.5pt]
    \small
    Carefully analyze the following fallacious argument: \\
    Premise 1: ``\texttt{<$p_0$>}'' \\
    Premise 2: ``\texttt{<$s_i$>}'' \\
    Premise 3: ``'' \\
    Therefore: ``\texttt{<claim>}'' \\

Both Premise 1 and Premise 2 originate from a reputable scientific document. The claim is deduced from information presented in Premise 1. However, Premise 2 introduces doubt, suggesting that the claim is an invalid conclusion based on the scientific document. \\

Your objective is to precisely identify and articulate the fallacious reasoning in Premise 3 (the fallacious premise). \textbf{This reasoning must robustly support the claim, ensuring that Premise 2 does not undermine the claim as a valid conclusion.} Consider only fallacies from the provided fallacy inventory. Present each fallacious premise alongside the applied fallacy class in this format:
\\

    Fallacious Premise: <fallacious premise>; Applied Fallacy Class: <applied fallacy class>.
    \\

If multiple fallacies are applicable, list them in order of relevance.

    \end{tcolorbox}
    \caption{\textbf{P5 Auto:} We automatically optimized the \emph{P2 Support} template by asking ChatGPT to improve the prompt for clarity and conciseness.}
    \label{fig:prompt-reconstruct-auto}
    \end{figure*}

        \begin{figure*}[!ht]
    \begin{tcolorbox}[left=1pt, right=1pt, top=1pt, bottom=1pt, boxrule=0.5pt]
    \small
Carefully analyze the following fallacious argument: \\
    Premise 1: ``\texttt{<$p_0$>}'' \\
    Premise 2: ``\texttt{<$s_i$>}'' \\
    Premise 3: ``'' \\
    Therefore: ``\texttt{<claim>}'' \\

Both Premise 1 and Premise 2 originate from a reputable scientific document.
The claim is deduced from information presented in Premise 1.
However, Premise 2 introduces doubt, suggesting that the claim is an invalid conclusion based on the scientific document.
 \\
 
Your objective is to precisely identify and articulate the fallacious reasoning in Premise 3 (the fallacious premise).
Do not repeat the claim itself, Premise 1, or Premise 2 when generating the fallacious Premise 3. \textbf{Make sure the generated Premise 3 connects Premise 1 and Premise 2 to robustly support the claim, and ensure that Premise 2 does not undermine the claim as a valid conclusion.}
Present each fallacious premise alongside the applied fallacy class in this format:
\\

    Fallacious Premise: <fallacious premise>; Applied Fallacy Class: <applied fallacy class>. \\

If multiple fallacies are applicable, list them in order of relevance.
    \end{tcolorbox}
    \caption{\textbf{P6 Auto-Connect:} An extension of \emph{P5 Auto} that explicitly requires the model to connect Premises 1 and 2 with the conclusion via the generated fallacious premise (as in \emph{P4 Connect}).}
    \label{fig:prompt-reconstruct-auto-connect}
    \end{figure*}

\subsubsection{Prompt for Consistency}
When evaluation the internal consistency of LLMs we use the prompt in Figure~\ref{fig:prompt-consistency}. As Premise 3 we insert the LLMs generated premise (Table~\ref{tab:subtask3-contextwise-generation}) or the annotated gold fallacious premise (Table~\ref{tab:subtask3-fallacywise-ablation}).
We measuring the consistency of the LLM we always provide the same level of detail about the fallacies (\emph{definition}, \emph{logical form}, \emph{example}) that were also available in the prompt when generating the fallacious premise.

\begin{figure}[!ht]
    \begin{tcolorbox}[left=1pt, right=1pt, top=1pt, bottom=1pt, boxrule=0.5pt]
    \small
Given the following argument and \texttt{<definitions with their logical forms and examples>}, determine which of the fallacies defined below occurs in Premise 3 of the provided argument.
The argument may contain multiple fallacies. Only detect the most fitting fallacy within Premise 3.
Explain your decision and conclude with the applied fallacy in a separate line at the end as "Fallacy: <fallacy class>". \\

\textbf{Fallacies:} \\
\texttt{<Fallacies with definition and/or logical form and/or example>}\\

\textbf{Argument:} \\
Premise 1: ``\texttt{<$p_0$>}'' \\
Premise 2: ``\texttt{<$s_i$>}'' \\
Premise 3: ``\texttt{<$\overline{p}_i$>}'' \\
Therefore: ``\texttt{<claim>}'' \\
    \end{tcolorbox}
    \caption{\textbf{Prompt Template for Consistency:} Template used to assess the consistency of the LLMs.}
    \label{fig:prompt-consistency}
    \end{figure}

\subsubsection{Prompt for Fallacy Classification without Premise}
\label{appendix:prompt-fallacy-classification-implicit}
We adapt the prompts from §\ref{appendix:prompts-reconstruction} to only instruct LLMs to classify the fallacy, but not generate the missing fallacious premise by adapting the task instructions as depicted in Figure~\ref{fig:prompt-fallacy-only-implicit}.

\begin{figure}[!ht]
    \begin{tcolorbox}[left=1pt, right=1pt, top=1pt, bottom=1pt, boxrule=0.5pt]
    \small
    \textbf{Task}: \\
Examine the following fallacious argument:\\

Premise 1: ``\texttt{<$p_0$>}'' \\
    Premise 2: ``\texttt{<$s_i$>}'' \\
    Premise 3: ``'' \\
    Therefore: ``\texttt{<claim>}'' \\

Premises 1 and 2 are sourced from the same credible scientific document.
The claim is based on the information in Premise 1.
However, Premise 2 suggests that the claim is an invalid conclusion from the scientific document. \\

A fallacy must be applied when connecting Premise 1 and Premise 2 to robustly support the claim.
Your task is to identify the applied fallacy class. Only consider fallacies from the provided fallacy inventory. \\

Present each applied fallacy class in this format: \\

    Fallacy: <applied fallacy class>. \\

If there are multiple applicable fallacies, list them in order of relevance. \\
    
    \end{tcolorbox}
    \caption{\textbf{Fallacy classification only:} Task instructions when the LLM is only tasked to classify the applied fallacy (without generating the fallacious premise).}
    \label{fig:prompt-fallacy-only-implicit}
    \end{figure}

\subsection{Paper Writing}
We used ChatGPT as assistant when condensing our paper content. We thoroughly reviewed and adjusted the paraphrased text for accuracy.

\end{document}